\documentclass[journal]{IEEEtran}%duoble
\newcommand{\ignore}[1]{}
\usepackage[utf8]{inputenc}
\usepackage[T1]{fontenc}
\usepackage{amsmath,epsfig}
\usepackage{graphicx}
\usepackage{bm}
\usepackage{amssymb}
\usepackage{cite}
\usepackage{array}
\usepackage{extarrows}
\usepackage{url}
\usepackage{float}

\usepackage{color}

\usepackage{multirow}
\usepackage{booktabs}
\usepackage{blkarray}
\usepackage{multirow}
\usepackage{array}
\usepackage{color}
\usepackage{pmat}
\begin{document}
%
% paper title
% can use linebreaks \\ within to get better formatting as desired
\title{Deeply-Coupled Convolution-Transformer with Spatial-temporal Complementary Learning for Video-based Person Re-identification}
\author{Xuehu~Liu,
        Chenyang~Yu,
        Pingping~Zhang$^{*}$, \textit{Member}, \textit{IEEE},
        and~Huchuan~Lu, \textit{Senior Member}, \textit{IEEE}
\thanks{
Copyright (c) 2023 IEEE. Personal use of this material is permitted. However, permission to use this material for any other purposes must  be obtained from the IEEE by sending an email to \textcolor{blue}{\underline{pubs-permissions@ieee.org}}.

%This work was supported in part by the National Natural Science Foundation of China under Grant No.62101092, 62293542, U1903215, the CAAI-Huawei MindSpore Open Fund under Grant CAAIXSJLJJ-2021-067A, the National Key R&D Program of China under Grant No.2018AAA0102001 and the Fundamental Research Funds for the Central Universities No.DUT22ZD210.

($^{*}$Corresponding author: Pingping Zhang.) % and Huchuan Lu

XH. Liu, CY. Yu and HC. Lu are with School of Information and Communication Engineering, Dalian University of Technology, Dalian, 116024, China.  (Email: {snowtiger, yuchenyang}@mail.dlut.edu.cn; lhchuan@dlut.edu.cn)

PP. Zhang is with School of Artificial Intelligence, Dalian University of Technology, Dalian, 116024, China. (Email: zhpp@dlut.edu.cn)
}
}
% make the title area
\maketitle
% The paper headers
\markboth{IEEE Transactions on Neural Networks and Learning Systems}{}
\begin{abstract}
Advanced deep Convolutional Neural Networks (CNNs) have shown great success in video-based person Re-Identification (Re-ID).
However, they usually focus on the most obvious regions of persons with a limited global representation ability.
Recently, it witnesses that Transformers explore the inter-patch relations with global observations for performance improvements.
In this work, we take both sides and propose a novel spatial-temporal complementary learning framework named Deeply-Coupled Convolution-Transformer (DCCT) for high-performance video-based person Re-ID.
Firstly, we couple CNNs and Transformers to extract two kinds of visual features and experimentally verify their complementarity.
Further, in spatial, we propose a Complementary Content Attention (CCA) to take advantages of the coupled structure and guide independent features for spatial complementary learning.
In temporal, a Hierarchical Temporal Aggregation (HTA) is proposed to progressively capture the inter-frame dependencies and encode temporal information.
Besides, a gated attention is utilized to deliver aggregated temporal information into the CNN and Transformer branches for temporal complementary learning.
Finally, we introduce a self-distillation training strategy to transfer the superior spatial-temporal knowledge to backbone networks for higher accuracy and more efficiency.
In this way, two kinds of typical features from same videos are integrated mechanically for more informative representations.
Extensive experiments on four public Re-ID benchmarks demonstrate that our framework could attain better performances than most state-of-the-art methods.
\end{abstract}
\IEEEpeerreviewmaketitle
\section{Introduction}

\IEEEPARstart{V}ideo-based person Re-Identification (Re-ID) aims to retrieve the corresponding videos of given pedestrians across cameras and times.
This advanced technology is beneficial to building intelligent urban for scenario surveillance and crime investigation.
Generally, the spatial appearance of persons changes dynamically in videos, leading to great challenges for re-identifying the same person.
In addition, the temporal relationships among multiple frames are worth being exploited to encode motion information.
Thus, how to obtain robust spatial-temporal representations of videos is widely seen as the key to tackle video-based person Re-ID.
%------------
\begin{figure}
\centering
\resizebox{0.48\textwidth}{!}
{
\begin{tabular}{@{}c@{}c@{}}
\includegraphics[width=0.35\linewidth,height=0.45\linewidth]{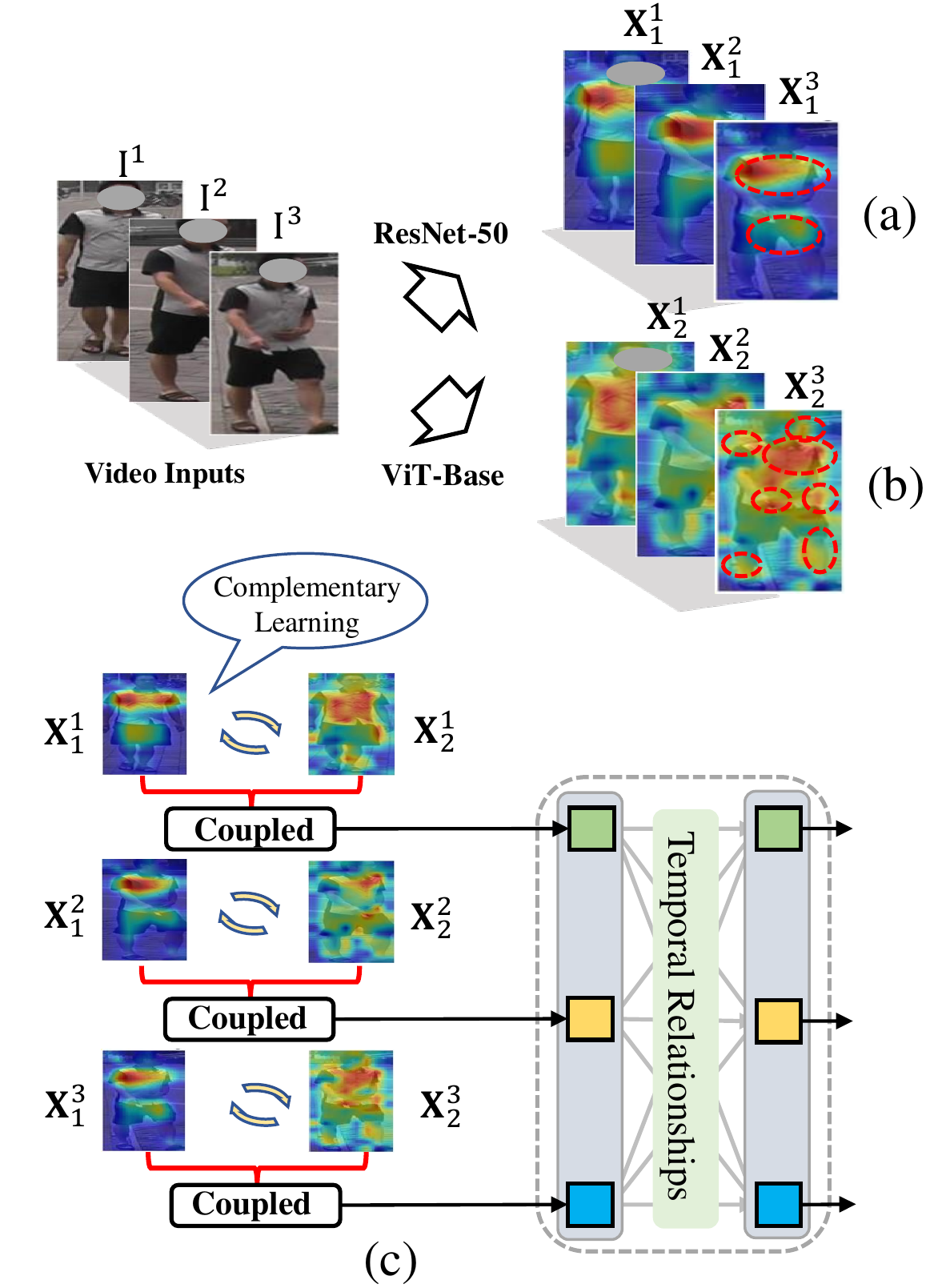} \\
\end{tabular}
}
%\vspace{-2mm}
\caption{Insights of our proposed method. (a) Feature maps of ResNet-50~\cite{he2016deep}, (b) Feature maps of ViT-Base~\cite{dosovitskiy2020image}. (c) Illustration of our proposed paradigm.
}
\vspace{-4mm}
\label{fig:motivation}
\end{figure}
%-------------------------------------------------------------

In previous methods, deep Convolutional Neural Networks (CNNs) are popular to obtain discriminative features for object Re-ID~\cite{cheng2016person, wang2021pyramid, ye2021dynamic, wu2021person, wang2022Pro}.
They exhibit impressive capabilities in local feature extraction.
As shown in Fig.~\ref{fig:motivation}(a), the main parts of pedestrians are covered with high brightness, which are discriminative cues for person retrieval.
However, naive CNNs usually focus on these obvious regions of persons with a limited global perception.
Recently, Transformers~\cite{dosovitskiy2020image,liu2021video} are rapidly developing in the computer vision field.
They generate local patches and model the inter-patch relations to obtain global observations.
Unlike typical CNNs, Transformers highlight more regions in spatial, as shown in Fig.~\ref{fig:motivation}(b).
They are helpful to identify pedestrians.
The key reason of their difference may localize in the basic operations of deep feature extraction in CNNs and Transformers.
In fact, the convolution kernels of CNNs have limited sizes, tending to recognize the most relevant patterns of objects.
While the self-attention blocks in Transformers capture the relational dependencies among contextual information, which helps to obtain integral representations from the global perception.

In light of the key differences, some methods combine CNNs and Transformers to encode local and global visual information for stronger representations.
For example, works in~\cite{xiao2021early,d2021convit} replace the patch stem with convolutions to improve the local encoding ability of Transformers.
Further, other works~\cite{liu2021video,zhang2021spatiotemporal,he2021dense} utilize excellent CNNs, such as ResNet-50~\cite{he2016deep}, to extract local features, following several Transformer layers to obtain final global representations.
Although effective, the simple combination cannot fully take advantages of CNNs and Transformers to generate complementary representations.
Thus, Peng~\emph{et al.}~\cite{peng2021conformer} combine ResNet-50 and ViT~\cite{dosovitskiy2020image} in parallel to obtain different representations and interact their spatial information with layer-by-layer.
It indeed attains impressive performance while consuming a huge amount of computational cost due to their frequent interactions between CNNs and Transformers.

In this work, we propose a novel complementary learning framework named Deeply-Coupled Convolution-Transformer (DCCT) for video-based person Re-ID.
Typically, in video-based person Re-ID task, spatial feature extraction and temporal representation learning are widely considered as two essential components.
Thus, in this work, we consider the complementary learning between CNNs and Transformers in both spatial and temporal.
In spatial feature extraction, we design a Complementary Content Attention (CCA) module to achieve spatial complementary learning.
It is deployed at the end of CNNs and Transformers for more efficiency.
In temporal representation learning, we propose a Hierarchical Temporal Aggregation (HTA) module to progressively encode and integrate two kinds of temporal features.
Besides, a gated attention is designed to deliver aggregated video information to two branches for temporal complementary learning.
Finally, in order to better optimize our framework, a self-distillation training strategy is proposed, which transfers the video-level knowledge into each frame-level feature of CNNs and Transformers.
With the self-distillation, we not only utilize the aggregated spatial-temporal information to guide the features of backbones for higher accuracy, but also reduce the inference time for more efficiency.
Extensive experiments on four public Re-ID benchmarks demonstrate that our framework could attain better performance than most state-of-the-art methods.

In summary, our contributions are four folds:
\begin{itemize}
\item We experimentally verify the complementarity between deep CNNs and Transformers, and propose a novel spatial-temporal complementary learning framework (\emph{i.e.}, DCCT) for video-based person Re-ID.
\item We propose a Complementary Content Attention (CCA) to enhance spatial features and a Hierarchical Temporal Aggregation (HTA) to integrate temporal information. They are designed for complementary learning and are effective in extracting robust representations.
\item We introduce a self-distillation training strategy to transfer video-level knowledge into frame-level features for higher accuracy and more efficiency.
\item Extensive experiments on four public Re-ID benchmarks demonstrate that our framework attains better performance than most state-of-the-art methods.
\end{itemize}

The rest of this work is organized as follows:
Sec.~II describes the related works of our method.
Sec.~III elaborates our approach, including the introduction of CCA, HTA and the self-distillation training strategy.
Sec.~IV presents the datasets, experimental settings and experimental results.
Sec.~V draws the conclusion and presents the future works.
%-------------------------------------------------------------------------
\section{Related Work}
\subsection{Spatial-temporal Learning}
Different from image-based tasks~\cite{li2018deep, li2021ctnet, jin2020deep, li2020weakly, ye2020augmentation}, the video-based tasks take videos as inputs, which contain richer spatial-temporal information than static images.
Thus, how to explore spatial-temporal cues is the key to tackle video-based person Re-ID.
On the one hand, to extract robust spatial features, Fu \emph{et al.}~\cite{fu2019sta} design a soft-attention module to weight horizontal parts of persons.
Gu~\emph{et al.} ~\cite{gu2020appearance} introduce an appearance-preserving module to original 3D convolutions and address the appearance destruction problem.
Zhang~\emph{et al.}~\cite{zhang2020multi} utilize a set of representative references as global guidance for multi-granularity semantics.
Chen~\emph{et al.}~\cite{chen2020temporal} disentangle video features into temporal coherent features and dynamic features, then highlight them by adversarial augmented temporal motion noise.
Bai \emph{et al.}~\cite{bai2022salient} design a salient-to-broad module to gradually enlarge the attention regions.
On the other hand, to improve the temporal representation ability, Liu~\emph{et al.}~\cite{liu2021watching} propose a temporal reciprocal learning mechanism to adaptively enhance or accumulate disentangled spatial features.
Li~\emph{et al.}~\cite{li2019multi} combine 2D CNNs and 3D CNNs to explicitly leverage spatial and temporal cues.
Hou~\emph{et al.}~\cite{hou2020temporal} design a temporal learning network to progressively discover diverse visual cues.
The aforementioned methods have shown great success in spatial or temporal learning for video-based person Re-ID.
However, these CNN-based methods have limited global representation capabilities and usually focus on salient local regions of persons.
We find that Transformers have strong characteristics in extracting global features.
Thus, different from previous works, we take the advantages of CNNs and Transformers, and propose a new spatial-temporal complementary learning framework for more informative video representations.
%-----------------------------------
\begin{figure*}[t]
\centering
\resizebox{0.95\textwidth}{!}
{
\begin{tabular}{@{}c@{}c@{}}
\includegraphics[width=1\linewidth,height=0.4\linewidth]{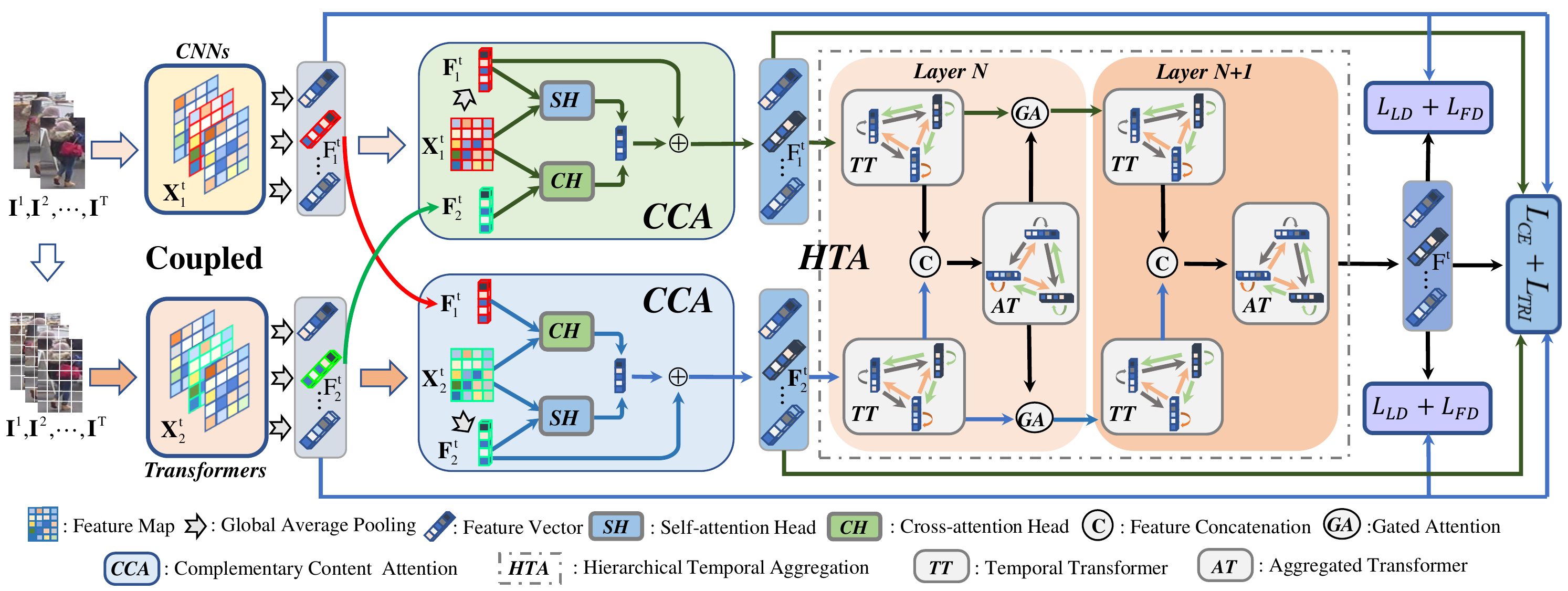} \\
\end{tabular}
}
%\vspace{-2mm}
\caption{The overall structure of our proposed framework.
Firstly, typical CNNs and Transformers are utilized to extract spatial feature maps of each frame.
Then, the Complementary Content Attention (CCA), including two Self-attention Heads (SH) and two Cross-attention Heads (CH), is introduced for spatial attention and complementary learning.
Afterwards, the Hierarchical Temporal Aggregation (HTA) is adopted to encode and integrate rich temporal information progressively.
To improve the ability, Temporal Transformers (TTs) are utilized for feature embedding, Aggregated Transformers (ATs) are utilized for feature aggregation, and Gated Attentions (GAs) are utilized for temporal complementary learning.
Finaly, the cross-entropy loss, triplet loss, logistic distillation and feature distillation are combined to train the whole framework.
}
\label{fig:Framework}
\vspace{-4mm}
\end{figure*}
%-------------------------------------------------------------
\subsection{Transformers for Person Re-identification}
Recently, Transformers have been applied to numerous computer vision tasks, such as image classification~\cite{dosovitskiy2020image,liu2021swin}, object detection~\cite{carion2020end}, semantic segmentation~\cite{wang2021end}, video inpainting~\cite{zeng2020learning} and so on.
Inspired by that, He~\emph{et al.}~\cite{he2021transreid} deploy ViT with a jigsaw patch module for object Re-ID.
Zhang~\emph{et al.}~\cite{zhang2021hat} design a hierarchical aggregation Transformer to aggregate multi-scale features with multi-granularity supervision for image-based person Re-ID.
In video-based person Re-ID, Zhang~\emph{et al.}~\cite{zhang2021spatiotemporal} utilize a spatial Transformer to encode image patches following a temporal Transformer to encode frame features.
He~\emph{et al.}~\cite{he2021dense} propose a hybrid framework with a dense interaction Transformer to model spatial-temporal inherence.
Liu~\emph{et al.}~\cite{liu2021video} construct a trigeminal Transformer in spatial, temporal and spatial-temporal views for comprehensive video representations.
All these methods have shown that Transformers can model the inter-patch relations for strong representations.
However, these methods do not take the full abilities of CNNs and Transformers in spatial and temporal learning, and lack direct complementary learning.
Different from existing Transformer-based Re-ID methods, our approach deeply couples CNNs and Transformers in parallel to obtain two kinds of visual representations and achieves complementary learning in both spatial and temporal.
\subsection{Knowledge Distillation}
Knowledge Distillation (KD)~\cite{hinton2015distilling} aims to transfer the knowledge from a larger and more accurate model (\emph{i.e.}, a teacher) to a more compact one (\emph{i.e.}, a student).
It has gained some attention in person Re-ID for its promise of more effective training~\cite{zhang2019your}, better accuracy~\cite{romero2014fitnets} and more efficiency~\cite{tung2019similarity,park2019relational}.
For example, Zheng~\emph{et al.}~\cite{zheng2021pose} design a simple KD framework to transfer the knowledge from the pose-guided branches to the main backbone branch during the training.
To improve the distillation, Zhang~\emph{et al.}~\cite{zhang2018deep} present a deep mutual learning strategy to learn collaboratively and teach each other.
Ge~\emph{et al.}~\cite{ge2020mutual} improve the discriminative capability of Re-ID models by integrating multiple student sub-networks in a mutual learning manner.
Zhou~\emph{et al.}~\cite{zhou2019discriminative} propose a layer-wise KD framework, which regards the high-level features of the same network as the teacher and transfers knowledge to the low-level features.
Different from these methods, in this work, we introduce the self-distillation training strategy that takes the aggregated video representation as a teacher and transfer its knowledge to the frame-level features of CNNs and Transformers.
In this way, our framework can be distilled in a top-down manner, gaining higher accuracy and more efficiency.
%-----------------------------------------------
\begin{table}[]
\caption{Symbols and representations. Subscript 1 means it is from the CNN branch, while subscript 2 means it is from the Transformer branch.}
\label{tab:Symbols and Representations}
\resizebox{0.48\textwidth}{!}{%
\begin{tabular}{c|ll}
	\cline{1-2}
	Symbols                                                                           & Representations                                                                                                           &  \\ \cline{1-2}
	${I}^t$                                                                    & Image of $t$-th frame in a video.                                                                                           &  \\ \cline{1-2}
	\multirow{2}{*}{$\textbf{X}^t_1,  \textbf{X}^t_2$}                                & Feature maps after backbones.                                                                                             &  \\
	& $\mathbf{{X}}^{t}_{1} \in {\mathbb{R}}^{H\times W\times C_1},\mathbf{{X}}^{t}_{2} \in {\mathbb{R}}^{H\times W\times C_2}$. &  \\ \cline{1-2}
	\multirow{2}{*}{$\tilde{\emph{\textbf{X}}}^t_1,  \tilde{\emph{\textbf{X}}}^t_2$}                & Feature vectors after GAP from $\textbf{X}^t_1,  \textbf{X}^t_2$.                                                         &  \\
	& $\tilde{\emph{\textbf{X}}}^t_1 \in {\mathbb{R}}^{C_1}, \tilde{\emph{\textbf{X}}}^t_2 \in {\mathbb{R}}^{C_2}$.                            &  \\ \cline{1-2}
	$\emph{\textbf{F}}^t_1, \emph{\textbf{F}}^t_2$                                                  & Feature vectors after CCA. $\emph{\textbf{F}}^t_1 \in {\mathbb{R}}^{C_1}, \emph{\textbf{F}}^t_2 \in {\mathbb{R}}^{C_2}$.                 &  \\ \cline{1-2}
	$\emph{\textbf{S}}^t_1, \emph{\textbf{S}}^t_2$                                                  & Feature vectors after TT. $\emph{\textbf{S}}^t_1 \in {\mathbb{R}}^{C_1}, \emph{\textbf{S}}^t_2 \in {\mathbb{R}}^{C_2}$.                  &  \\ \cline{1-2}
	$\emph{\textbf{S}}_3$                                                                    & Feature vector after AT and TAP. $\emph{\textbf{S}}_3 \in {\mathbb{R}}^{C_1 + C_2}$.                                                     &  \\ \cline{1-2}
	$\textbf{\emph{M}}^t_1, \textbf{\emph{M}}^t_2$                                                  & Feature vectors after GA. $\emph{\textbf{M}}^t_1 \in {\mathbb{R}}^{C_1}, \emph{\textbf{M}}^t_2 \in {\mathbb{R}}^{C_2}$.                  &  \\ \cline{1-2}
	\multirow{2}{*}{$\textbf{A}^s_1, \textbf{A}^s_2, \textbf{A}^c_1, \textbf{A}^c_2$} & Attention maps from SH and CH.                                                                                            &  \\
	& $\textbf{A}^s_1, \textbf{A}^s_2, \textbf{A}^c_1, \textbf{A}^c_2 \in {\mathbb{R}}^{H \times W}$.                            &  \\ \cline{1-2}
	$\emph{\textbf{A}}^g_1, \emph{\textbf{A}}^g_2$                                                  & Attention vectors from GA. $\emph{\textbf{A}}^g_1 \in {\mathbb{R}}^{C_1}, \emph{\textbf{A}}^g_2  \in {\mathbb{R}}^{C_2}$.                &  \\ \cline{1-2}
\end{tabular}%
}
\vspace{-4mm}
\end{table}

\section{Proposed Method}
In this section, we introduce the proposed spatial-temporal complementary learning framework to obtain video representations.
The overall framework is shown in Fig.~\ref{fig:Framework}.
It contains four main components: Coupled Convolution-Transformer, Complementary Content Attention (CCA) for spatial complementary attention, Hierarchical Temporal Aggregation (HTA) for temporal complementary aggregation, and self-distillation training strategy.
We elaborate these key components in the following subsections.
For clarification, the typical symbols and representations used in this work are listed in Tab.~\ref{tab:Symbols and Representations}.

\subsection{Coupled Convolution-Transformer}
As stated in Sec.~I, the convolution kernels of CNNs tend to recognize the most relevant patterns of objects in local regions, while the self-attention blocks in Transformers capture the relational dependencies of contextual information to obtain the global perception.
In order to take advantage of their unique strengths, we design a dual-path structure to extract two kinds of features from CNNs and Transformers.
Specifically, given an image sequence ${\{{\emph{I}}^t \}}^T_{t=1}$ of one person, the CNN (\emph{e.g.}, ResNet-50~\cite{he2016deep}) and Transformer (\emph{e.g.}, ViT-Base~\cite{dosovitskiy2020image}) backbones are utilized to obtain frame-level convolutional feature maps ${\{\textbf{X}^t_1 \}}^T_{t=1}$ and Transformer-based feature maps ${\{\textbf{X}^t_2 \}}^T_{t=1}$, respectively.
Here, $\textbf{X}^t_1 \in {\mathbb{R}}^{H \times W \times C_1 }$, $\textbf{X}^t_2 \in {\mathbb{R}}^{H \times W \times C_2}$,
$H$ and $W$ represent the height and width of spatial feature maps, $C_1$ and $C_2$ are the number of channels.
In our baseline method, these two kinds of frame-level spatial features are followed by Global Average Pooling (GAP) and Temporal Average Pooling (TAP) to obtain global features $\tilde{\emph{\textbf{X}}}_1$ and $\tilde{\emph{\textbf{X}}}_2$.
After that, $\tilde{\emph{\textbf{X}}}_1$ and $\tilde{\emph{\textbf{X}}}_2$ are concatenated as the final video representation for person retrieval.
In the training stage, the cross-entropy loss and triplet loss are utilized to optimize the framework.
In person Re-ID, our \emph{Coupled Convolution-Transformer} can be seen as a new paradigm for deep feature extraction.
Based on this, we further propose a new complementary learning framework in both spatial and temporal for video-based person Re-ID.
\subsection{Complementary Content Attention}
\begin{figure}[t]
\centering
\resizebox{0.48\textwidth}{!}
{
\begin{tabular}{@{}c@{}c@{}}
\includegraphics[width=0.5\linewidth,height=0.36\linewidth]{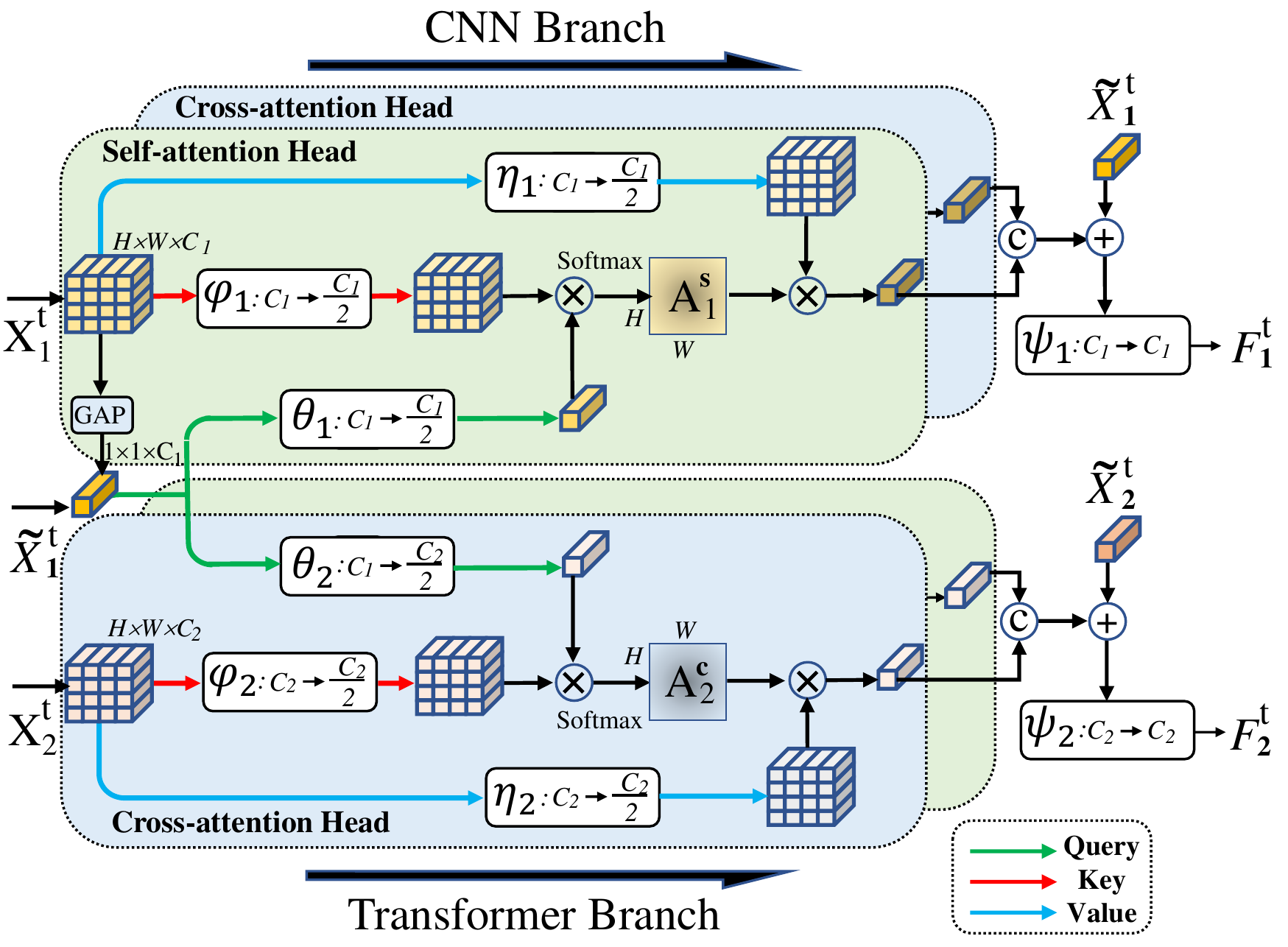} \\
\end{tabular}
}
%\vspace{-2mm}
\caption{The proposed Complementary Content Attention (CCA) with Self-attention Heads (SH) and Cross-attention Heads (CH).
}
\label{fig:CCA}
\vspace{-4mm}
\end{figure}

With the coupled Convolution-Transformer, two kinds of spatial features ${\{\textbf{X}^t_1 \}}^T_{t=1}$ and ${\{\textbf{X}^t_2 \}}^T_{t=1}$ have been extracted as complementary representations.
Beyond direct concatenation, we propose a Content Complementary Attention (CCA) module to effectively exploit these representations for spatial attention and complementary learning.
Our CCA comprises the CNN branch and the Transformer branch, as shown in Fig.~\ref{fig:CCA}.
Each branch contains a Self-attention Head (SH) and a Cross-attention Head (CH).
In SH, the global feature from one branch attends to its local features for self attention, which helps to focus on the spatially obvious cues of objects.
In CH, the global feature from one branch attends to the local features from another branch for cross attention, which helps to capture more complementary information in spatial.
Specifically, taking the $t$-th frame as an example, in the SH of the CNN branch, the local feature $\mathbf{{X}}^{t}_{1}$
is seen as the \emph{key}, the global feature $\tilde{\emph{\textbf{X}}}^t_{1} \in {\mathbb{R}}^{C_1}$ is seen as the \emph{query}.
The \emph{query} and \emph{key} are feed into two non-shared linear projections $\theta_1$ and ${\varphi}_1$, respectively.
Then, the self-attention map $\textbf{A}^s_1 \in {\mathbb{R}}^{H\times W}$ is estimated by
\begin{equation}\label{2}
\textbf{A}^s_1 =  \sigma(\mathbf{\varphi}_1 (\mathbf{{X}}^t_1) \otimes \mathbf{\theta}_1 (\tilde{\textbf{\emph{X}}}^t_{1})),
\end{equation}
where $\sigma$ is the softmax activation function,
$\otimes$ indicates the matrix multiplication.
With the reference of the global feature, the discriminative regions of objects are highlighted.

Meanwhile, in the CH of the Transformer branch, the global feature $\tilde{\textbf{\emph{X}}}^t_{1} \in {\mathbb{R}}^{C_1}$ from CNN is the \emph{query}, the local feature $\mathbf{{X}}^{t}_{2}$ is the \emph{key}.
Similarly, two linear projections $\theta_2$ and ${\varphi}_2$ are applied to the \emph{query} and \emph{key}.
Then, we estimate the cross-attention map $\textbf{A}^c_2 \in {\mathbb{R}}^{H\times W}$, as follows
\begin{equation}\label{4}
\textbf{A}^c_2 = \sigma(\mathbf{\varphi}_2 (\mathbf{{X}}^t_2) \otimes \mathbf{\theta}_2 (\tilde{\textbf{\emph{X}}}^t_{1})).
\end{equation}
It is worth noting that our CHs can implement the spatial complementary learning with the cross-branch mutual guidance.
Due to the distinctness between CNN and Transformer, the \emph{query} features contain different semantic information.
Typically, the features from CNN are more concerned with local salient information, while the features from Transformer are more concerned with global holistic information.
As a result, the convolutional features facilitate Transformer-based features to reduce the impact of noises in the background, and Transformer-based features assist convolutional features to capture more meaningful cues.

In the same way, the self-attention map $\textbf{A}^s_2$ and the cross-attention map $\textbf{A}^c_1$ are generated from the Transformer branch and CNN branch.
The visualizations of attention maps are shown in Fig.~\ref{fig:visualization_CCA}.
%-------------------------------------------------------------
%
After that, the original local features seen as the \emph{values} are multiplied with attention maps to generate attentive features.
Then, the self-attentive feature and the cross-attentive feature are concatenated in channel and added with the original global features for residual connections,
\begin{equation}\label{5}
\mathbf{{\textbf{\emph{F}}}}^t_1 = \mathbf{\psi}_1 ([\textbf{A}^s_1 \otimes \mathbf{\eta}_1 (\mathbf{{X}}^t_1), \textbf{A}^c_1 \otimes \mathbf{\tilde{\eta}}_1 (\mathbf{{X}}^t_1)] + \tilde{\textbf{\emph{X}}}^t_{1}),
\end{equation}
\begin{equation}\label{6}
\mathbf{{\textbf{\emph{F}}}}^t_2 = \mathbf{\psi}_2 ( [\textbf{A}^s_2 \otimes \mathbf{\tilde{\eta}}_2 (\mathbf{{X}}^t_2), \textbf{A}^c_2 \otimes  \mathbf{\eta}_2 (\mathbf{{X}}^t_2)] + \tilde{\textbf{\emph{X}}}^t_{2}),
\end{equation}
where $\textbf{\emph{F}}^t_1 \in {\mathbb{R}}^{C_1}$ and $\textbf{\emph{F}}^t_2 \in {\mathbb{R}}^{C_2}$,
[,] represents the feature concatenation in the channel,
$\mathbf{\eta}_1, \mathbf{\eta}_2, \mathbf{\tilde{\eta}}_1, \mathbf{\tilde{\eta}}_2$ are four linear projections, $\mathbf{\psi}_1$ and $\mathbf{\psi}_2$ are two fully-connected layers.
In this way, the enhanced convolutional feature ${\{\textbf{\emph{F}}^t_1 \}}^T_{t=1}$ and Transformer-based features ${\{\textbf{\emph{F}}^t_2 \}}^T_{t=1}$ are obtained from the CNN and Transformer branches.
Thus, our CCA can perform spatial attention and achieve complementary learning to get better frame-level representations.
%------------------------------------------------
\begin{figure}[t]
\centering
\resizebox{0.48\textwidth}{!}
{
\begin{tabular}{@{}c@{}c@{}}
\includegraphics[width=1.0\linewidth,height=0.55\linewidth]{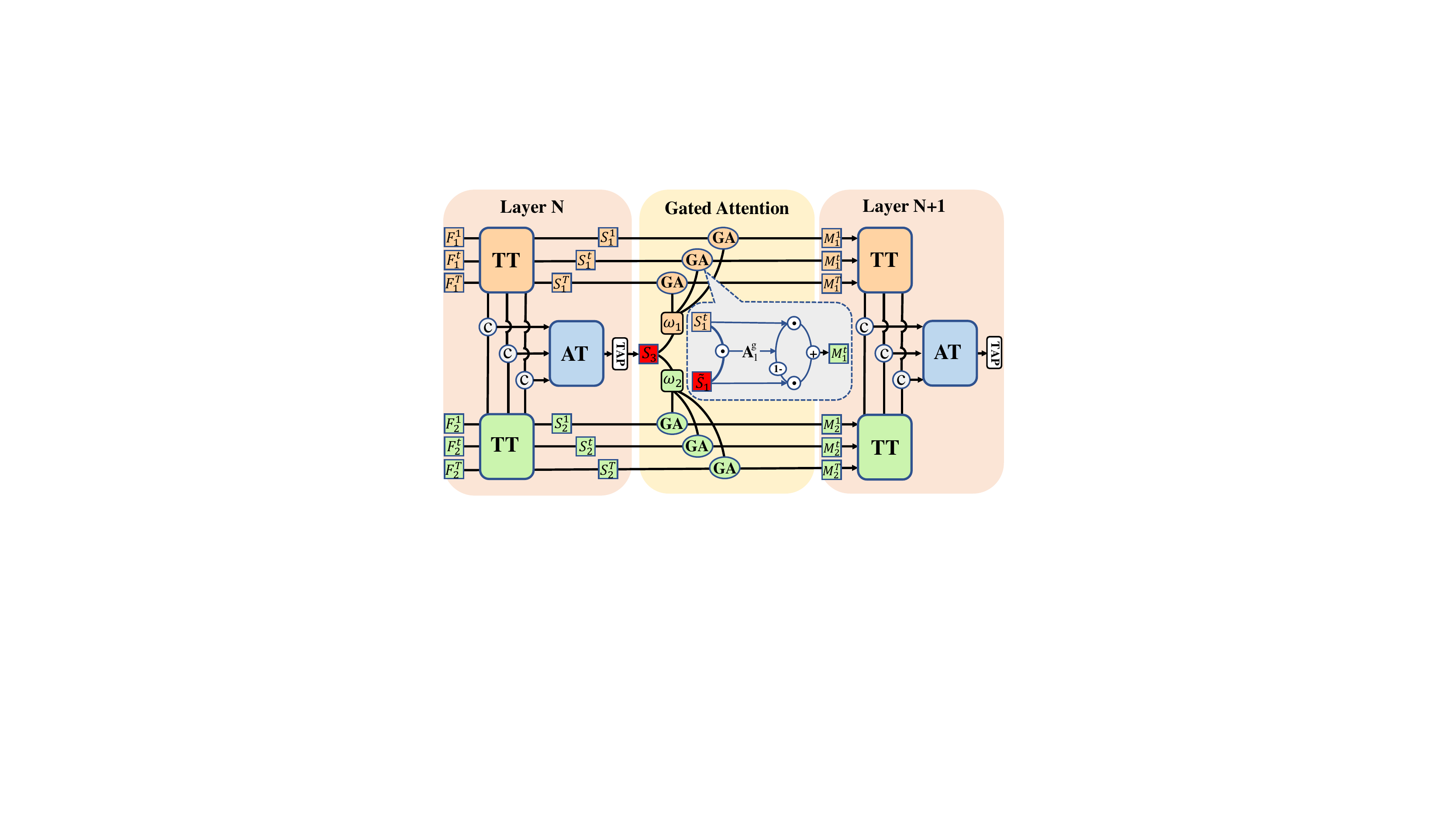} \\
\end{tabular}
}
%\vspace{-2mm}
\caption{The proposed Hierarchical Temporal Aggregation (HTA).
}
\label{fig:HTA}
\vspace{-4mm}
\end{figure}
%------------------------------------------------------------
\subsection{Hierarchical Temporal Aggregation}
After spatial complementary learning by CCA, subsequent temporal learning should be considered to obtain the final video representations.
To this end, we propose a Hierarchical Temporal Aggregation (HTA) to encode and aggregate multi-frame features in temporal.
The HTA is shown in Fig.~\ref{fig:HTA}, which has a hierarchical structure.
Each layer of HTA contains three parts: two Temporal Transformers (TT), an Aggregated Transformer (AT), and a Gated Attention (GA).
In TTs, we capture the inter-frame relations to encode two kinds of temporal features $ \textbf{F}_1 ({\{\textbf{\emph{F}}^t_1 \}}^T_{t=1})$ and $\textbf{F}_2 ({\{\textbf{\emph{F}}^t_1 \}}^T_{t=1})$, respectively.
In AT, two kinds of features are concatenated in channel for further temporal encoding.
Followed by a TAP, the aggregated video-level feature are obtained.
Meanwhile, a GA is designed to deliver the aggregated information to the CNN and Transformer branches for temporal complementary learning.

Specifically, taking one TT as an example, the temporal feature $\textbf{F}_1$ is added with a learnable position embedding~\cite{dosovitskiy2020image}, then passed into a Multi-Head Self-Attention (MHSA) layer.
In each head of MHSA, the self-attention is defined as
\begin{equation}\label{7}
\textbf{Z}_h = \sigma(\frac{\textbf{Q}\textbf{K}^{\top}}{\sqrt{d}})\textbf{V},
\end{equation}
where $(\cdot)^{\top}$ indicates the matrix transpose operation.
$\textbf{Q}$, $\textbf{K}$, $\textbf{V}$ $\in \mathbb{R}^{T\times d}$ are generated from $\textbf{F}_1$ by three linear projections.
$d= \frac{C_1}{N_h}$ and $N_h$ is the number of heads.
Then, the outputs of multiple heads are concatenated in channel to generate the feature $\textbf{Z}=[\textbf{Z}_1, \cdots, \textbf{Z}_{N_h}]$.
After that, a frame-wise feed-forward network $\mathbf{\Phi}$ is applied for further feature embedding,
\begin{equation}\label{8}
\textbf{S}_1 = \mathbf{\Phi} (\textbf{Z} + \textbf{F}_1),
\end{equation}
where $\mathbf{\Phi}$ includes two fully-connected layer with a ReLU activation function and a normalization layer.
Similarly, the $\textbf{S}_2$ can be obtained from another TT.
After TTs, two kinds of features $\textbf{S}_1$ and $\textbf{S}_2$ are concatenated in channel and passed into ATs for aggregated temporal learning.
In this way, we explore the inter-frame relationships and propagate information across frames for temporal representation learning.
Finally, a TAP is utilized to obtain the video-level feature $\textbf{\emph{S}}_3 \in \mathbb{R}^{C_1 + C_2}$.

To achieve complementary learning, we further design a GA to adaptively deliver the aggregated video information to the two temporal features.
In GA, we first decouple $\textbf{S}_3$ into $\tilde{\textbf{\textbf{S}}}_{1} \in \mathbb{R}^{C_1}$ and $\tilde{\textbf{\textbf{S}}}_{2} \in \mathbb{R}^{C_2}$ by two linear projections $\omega_1$ and $\omega_2$,
\begin{equation}\label{10}
\tilde{\textbf{\emph{S}}}_{1} = \mathbf{\omega}_{1} (\textbf{\emph{S}}_3),
\end{equation}
\begin{equation}\label{11}
\tilde{\textbf{\emph{S}}}_{2} = \mathbf{\omega}_{2} (\textbf{\emph{S}}_3).
\end{equation}
Afterward, at the $t$-th time step, we estimate the channel attentions to explore which complementary information should be learned.
These attentions are utilized as the gates to control information communication and generated by
\begin{equation}\label{12}
\textbf{A}^g_1 = \mathbf{\phi}_1 ( \textbf{\emph{S}}_1^t \odot \tilde{\textbf{\emph{S}}}_{1} ),
\end{equation}
\begin{equation}\label{13}
\textbf{A}^g_2 = \mathbf{\phi}_2 ( \textbf{\emph{S}}_2^t \odot \tilde{\textbf{\emph{S}}}_{2}  ),
\end{equation}
where $\mathbf{\phi}_{1}$ and $\mathbf{\phi}_{2}$ are two non-shared fully-connected layers followed by the sigmoid activation function.
Finally, the gated attention weights are utilized to aggregate two kinds of features for temporal complementary learning,
\begin{equation}\label{14}
\textbf{\emph{M}}_1^t = \textbf{A}^g_1 \odot \textbf{\emph{S}}_1^t  + ( 1 - \textbf{A}^g_1) \odot \tilde{\textbf{\emph{S}}}_{1},
\end{equation}
\begin{equation}\label{15}
\textbf{\emph{M}}_2^t = \textbf{A}^g_2 \odot \textbf{\emph{S}}_2^t  + ( 1 - \textbf{A}^g_2) \odot \tilde{\textbf{\emph{S}}}_{2}.
\end{equation}
In this way, GA can be utilized to control the the flow of information between two branches.
The aggregated video feature is adaptively selected in channels to update the original frame features to achieve temporal complementary learning.
Moreover, ${\{\textbf{\emph{M}}^t_1 \}}^T_{t=1}$ and ${\{\textbf{\emph{M}}^t_2 \}}^T_{t=1}$ are passed into the next layer of HTA.
With the stacked layers, temporal features are strengthened progressively.

In summary, our HTA explores the relationships among frames to progressively encode rich temporal information.
Besides, we aggregate convolutional features and Transformer-based features, and selectively deliver aggregated video features to each frame-level feature for temporal complementary learning.
Therefore, our HTA can encode and aggregate temporal information to generate better video representations.

\subsection{Self-distillation Training Strategy}
In this work, we adopt the cross-entropy loss~\cite{xiao2017joint} and triplet loss~\cite{ding2015deep} to train our framework.
As shown in Fig.~\ref{fig:Framework}, they are deployed at two stages: (1) Backbone supervision, (2) Final supervision.
Experimentally, if only employing the final supervision, the performances are very bad due to the weak supervision information.
To improve the learning ability, we introduce a new self-distillation strategy to train our framework for higher accuracy and more efficiency.
In the self-distillation, the aggregated feature is seen as a teacher and transfers its spatial-temporal knowledge to backbone networks.

Specifically, the self-distillation strategy consists of a logistic distillation and a feature distillation.
In the logistic distillation, the frame-level convolutional features $\{\tilde{\textbf{\emph{X}}}^t_{1} \}_{t=1}^{T}$ and Transformer-based features $\{\tilde{\textbf{\emph{X}}}_2^t \}_{t=1}^{T}$ are passed into two non-shared classifiers to generate the class probability of each identity,
\begin{equation}\label{16}
{p}^{i,t}_1 = \frac{\text{exp}(\textbf{W}^i_1 {\tilde{\textbf{\emph{X}}}}^t_1 )}{\sum_{j}^{I} \text{exp}(\textbf{W}^j_1 {\tilde{\textbf{\emph{X}}}}^t_1 ) },
\end{equation}
\begin{equation}\label{17}
{p}^{i,t}_2 = \frac{\text{exp}(\textbf{W}^i_2 {\tilde{\textbf{\emph{X}}}}^t_2 )}{\sum_{j}^{I} \text{exp}(\textbf{W}^j_2 {\tilde{\textbf{\emph{X}}}}^t_2 ) },
\end{equation}
where $\textbf{W}^i_1$ and $\textbf{W}^i_2$ are the weights of the $i$-th class in classifiers.
$I$ is the total number of classes.
We denote $\{{p}^{i,t}_1 \}^{I,T}_{i=1,t=1}$, $\{{p}^{i,t}_2 \}^{I,T}_{i=1,t=1}$, as $\{\textbf{P}^{t}_1 \}^{T}_{t=1}$, $\{\textbf{P}^{t}_2 \}^{T}_{t=1}$ respectively.
Besides, the aggregated video feature $\textbf{S}_3$ after CCA and HTA is utilized to compute its class probability,
\begin{equation}\label{18}
{p}^i_3 = \frac{\text{exp}(\textbf{W}^i_3 {\textbf{\emph{S}}}_3 )}{\sum_{j}^{I} \text{exp}(\textbf{W}^j_3 {\textbf{\emph{S}}}_3 ) },
\end{equation}
where $\textbf{W}^i_3$ is the weight of the $i$-th class in classifier.
$\{{p}^{i}_3 \}^{I}_{i=1}$ is denoted as $\textbf{P}_3$.
Afterwards, we utilize the Kullback-Leibler divergence~\cite{kullback1951on} to compute the logistic distillation loss,
\begin{equation}\label{18}
\mathcal{L}_{LD} = \sum_{t=1}^{T} \text{KL}(\textbf{P}_{1}^t, \textbf{P}_{3}) + \text{KL}(\textbf{P}_{2}^t, \textbf{P}_{3}).
\end{equation}
In the feature distillation, the video-level feature $\textbf{S}_3$ is seen as a teacher to directly guide frame-level convolutional features $\{\tilde{\textbf{X}}_1^t \}_{t=1}^{T}$ and Transformer-based features $\{\tilde{\textbf{X}}_2^t \}_{t=1}^{T}$.
Thus, the feature distillation loss can be computed by
\begin{equation}\label{18}
\mathcal{L}_{FD} = \sum_{t=1}^{T} \| \textbf{W}_1^h \tilde{\textbf{\emph{X}}}_1^t - \textbf{\emph{S}}_3\|_2^2 + \| \textbf{W}_2^h \tilde{\textbf{\emph{X}}}_2^t - \textbf{\emph{S}}_3\|_2^2,
\end{equation}
where $\textbf{W}_1^h$ and $\textbf{W}_2^h$ represent two hint networks, including bottleneck architectures to align the features on channel sizes.
Generally, video features have aggregated the spatial and temporal information from multiple frames.
Thus, compared with video-to-video or frame-to-frame distillations, our video-to-frame distillation is more suitable for both video and frame learning in video-based person Re-ID.

Overall, the proposed self-distillation training strategy is beneficial for higher accuracy and more efficiency.
For accuracy, the self-distillation could transfer spatial-temporal knowledge to the CNN or Transformer backbones and achieve higher accuracy.
For efficiency, after self-distillation training, one can remove CCA and HTA, and only use the features obtained from backbones for test.
It still attains better performance and reduces inference time.
Finally, all above losses are used to train the framework,
\begin{equation}\label{20}
\mathcal{L} = \lambda_1 \mathcal{L}_{CE} + \lambda_2 \mathcal{L}_{Tri} + \lambda_3 \mathcal{L}_{LD} + \lambda_4 \mathcal{L}_{FD}.
\end{equation}
As a result, the whole framework with deeply-supervised spatial-temporal complementary learning can be optimized.

\section{Experiments}
\subsection{Datasets and Evaluation Protocols}
In this paper, we conduct extensive experiments to evaluate our method on four public video-based person Re-ID datasets, \emph{i.e.}, MARS~\cite{zheng2016mars}, DukeMCMT-VID~\cite{zheng2017unlabeled}, iLIDS-VID~\cite{wang2014person} and
PRID-2011~\cite{hirzer2011person}.
MARS is a large-scale dataset, which consists of 1,261 identities around 18,000 video sequences.
On MARS, there are round 3,200 distractors sequences. Meanwhile, each sequence has 59 frames on average.
DukeMTMC-VID is another large-scale dataset.
It comprises 4,832 sequences from 1,812 identities including 408 distractor identities.
Each sequence has 168 frames on average.
iLIDS-VID is a small dataset collected by two cameras, which consists of 600 video sequences of 300 different identities.
PRID-2011 is another small dataset and consists of 400 image sequences for 200 identities from two cameras.
Following previous works, we adopt the Cumulative Matching Characteristic (CMC) and mean Average Precision (mAP)~\cite{zheng2016person} for evaluation.
\subsection{Experiment Settings}
Our experiments are implemented with two GTX V100 GPUs.
We adopt the Restricted Random Sampling (RRS)~\cite{li2018diversity} to generate sequential frames.
Each frame is augmented by random cropping, horizontal flipping, random erasing and resized to 256$\times$128.
The backbones of CNNs and Transformers are both pre-trained on the ImageNet dataset~\cite{deng2009imagenet}.
Besides, the batch-size is set to 16.
A mini-batch contains 8 person identifies, and each identify has 2 sequences.
The stochastic gradient descent~\cite{bottou2010large} algorithm is utilized to train our method with an initial learning rate $10^{-3}$, weight decay $5\times10^{-4}$ and nesterov momentum 0.9.
When training, the learning rate is decayed by 10 at every 15 epochs until the maximum 50 epochs.
Experimentally, the ${\lambda}_1, {\lambda}_2$ are set to 1, and the ${\lambda}_3, {\lambda}_4$ are set to 0.1.
We will release our code at https://github.com/flysnowtiger/DCCT.

%-------------------------------------------------------------------------
\begin{table}[t]
	\begin{center}
		\doublerulesep=0.5pt
		%\vspace{-2mm}
		\caption{Performance (\%) comparison on MARS~\cite{zheng2016mars} and DukeMTMC-VID~\cite{wu2018exploit} datasets.
			\textreferencemark indicates that Transformer is deployed after CNN.
			The methods can be divided into two groups, \emph{i.e.} CNN-based (C) and Transformer-based (T) methods.
			The texts in bold and underline highlight the best and second performances.}
		\label{table:mars duke sota}
		%\vspace{-2mm}
		\resizebox{0.45\textwidth}{!}
		{
			\begin{tabular}{c|llllllll|clllclll|clllclll}
				\hline
				\multicolumn{1}{l|}{\multirow{2}{*}{}} &
				\multicolumn{8}{l|}{} &
				\multicolumn{8}{c|}{MARS} &
				\multicolumn{8}{c}{DukeMCMT-VID} \\
				\multicolumn{1}{l|}{} &
				\multicolumn{8}{l|}{Methods} &
				\multicolumn{4}{c|}{mAP} &
				\multicolumn{4}{c|}{Rank-1} &
				\multicolumn{4}{c|}{mAP} &
				\multicolumn{4}{c}{Rank-1} \\ \hline
				\multirow{20}{*}{C} &
				\multicolumn{8}{l|}{EUG~\cite{wu2018exploit}} &
				\multicolumn{4}{c|}{67.4} &
				\multicolumn{4}{c|}{80.8} &
				\multicolumn{4}{c|}{78.3} &
				\multicolumn{4}{c}{83.6} \\
				&
				\multicolumn{8}{l|}{M3D~\cite{li2019multi}} &
				\multicolumn{4}{c|}{74.1} &
				\multicolumn{4}{c|}{84.4} &
				\multicolumn{4}{c|}{--} &
				\multicolumn{4}{c}{--} \\
				&
				\multicolumn{8}{l|}{STA~\cite{fu2019sta}} &
				\multicolumn{4}{c|}{80.8} &
				\multicolumn{4}{c|}{86.3} &
				\multicolumn{4}{c|}{94.9} &
				\multicolumn{4}{c}{96.2} \\
				&
				\multicolumn{8}{l|}{Attribute~\cite{zhao2019attribute}} &
				\multicolumn{4}{c|}{78.2} &
				\multicolumn{4}{c|}{87.0} &
				\multicolumn{4}{c|}{95.3} &
				\multicolumn{4}{c}{95.4} \\
				&
				\multicolumn{8}{l|}{VRSTC~\cite{hou2019vrstc}} &
				\multicolumn{4}{c|}{82.3} &
				\multicolumn{4}{c|}{88.5} &
				\multicolumn{4}{c|}{93.5} &
				\multicolumn{4}{c}{95.0} \\
				&
				\multicolumn{8}{l|}{GLTR~\cite{li2019global}} &
				\multicolumn{4}{c|}{78.5} &
				\multicolumn{4}{c|}{87.0} &
				\multicolumn{4}{c|}{93.7} &
				\multicolumn{4}{c}{96.3} \\
				&
				\multicolumn{8}{l|}{STE-NVAN~\cite{Liu2019SpatiallyAT}} &
				\multicolumn{4}{c|}{81.2} &
				\multicolumn{4}{c|}{88.9} &
				\multicolumn{4}{c|}{93.5} &
				\multicolumn{4}{c}{95.2} \\
				&
				\multicolumn{8}{l|}{NL-AP3D~\cite{gu2020appearance}} &
				\multicolumn{4}{c|}{85.6} &
				\multicolumn{4}{c|}{90.7} &
				\multicolumn{4}{c|}{96.1} &
				\multicolumn{4}{c}{97.2} \\
				&
				\multicolumn{8}{l|}{AFA~\cite{chen2020temporal}} &
				\multicolumn{4}{c|}{82.9} &
				\multicolumn{4}{c|}{90.2} &
				\multicolumn{4}{c|}{95.4} &
				\multicolumn{4}{c}{97.2} \\
				&
				\multicolumn{8}{l|}{TCL~\cite{hou2020temporal}} &
				\multicolumn{4}{c|}{85.1} &
				\multicolumn{4}{c|}{89.8} &
				\multicolumn{4}{c|}{96.2} &
				\multicolumn{4}{c}{96.9} \\
				&
				\multicolumn{8}{l|}{STGCN~\cite{yang2020spatial}} &
				\multicolumn{4}{c|}{83.7} &
				\multicolumn{4}{c|}{89.9} &
				\multicolumn{4}{c|}{95.7} &
				\multicolumn{4}{c}{97.9} \\
				&
				\multicolumn{8}{l|}{MGH~\cite{yan2020learning}} &
				\multicolumn{4}{c|}{85.8} &
				\multicolumn{4}{c|}{90.0} &
				\multicolumn{4}{c|}{--} &
				\multicolumn{4}{c}{--} \\
				&
				\multicolumn{8}{l|}{MGRA~\cite{zhang2020multi}} &
				\multicolumn{4}{c|}{85.9} &
				\multicolumn{4}{c|}{88.0} &
				\multicolumn{4}{c|}{--} &
				\multicolumn{4}{c}{--} \\
				&
				\multicolumn{8}{l|}{SSN3D~\cite{jiang2021ssn3d}} &
				\multicolumn{4}{c|}{86.2} &
				\multicolumn{4}{c|}{90.1} &
				\multicolumn{4}{c|}{96.3} &
				\multicolumn{4}{c}{96.8} \\
				&
				\multicolumn{8}{l|}{CTL~\cite{liu2021spatial}} &
				\multicolumn{4}{c|}{86.7} &
				\multicolumn{4}{c|}{91.4} &
				\multicolumn{4}{c|}{--} &
				\multicolumn{4}{c}{--} \\
				&
				\multicolumn{8}{l|}{GRL~\cite{liu2021watching}} &
				\multicolumn{4}{c|}{84.8} &
				\multicolumn{4}{c|}{91.0} &
				\multicolumn{4}{c|}{--} &
				\multicolumn{4}{c}{--} \\
				&
				\multicolumn{8}{l|}{BiCNet-TKS~\cite{hou2021bicnet}} &
				\multicolumn{4}{c|}{86.0} &
				\multicolumn{4}{c|}{90.2} &
				\multicolumn{4}{c|}{96.1} &
				\multicolumn{4}{c}{96.3} \\
				&
				\multicolumn{8}{l|}{STRF~\cite{aich2021spatio}} &
				\multicolumn{4}{c|}{86.1} &
				\multicolumn{4}{c|}{90.3} &
				\multicolumn{4}{c|}{96.4} &
				\multicolumn{4}{c}{97.4} \\
				&
				\multicolumn{8}{l|}{STMN~\cite{eom2021video}} &
				\multicolumn{4}{c|}{84.5} &
				\multicolumn{4}{c|}{90.5} &
				\multicolumn{4}{c|}{95.9} &
				\multicolumn{4}{c}{97.0} \\
				&
				\multicolumn{8}{l|}{PSTA~\cite{wang2021pyramid}} &
				\multicolumn{4}{c|}{85.8} &
				\multicolumn{4}{c|}{\underline{91.5}} &
				\multicolumn{4}{c|}{\underline{97.4}} &
				\multicolumn{4}{c}{\underline{98.3}} \\
				&
				\multicolumn{8}{l|}{SINet~\cite{bai2022salient}} &
				\multicolumn{4}{c|}{86.2} &
				\multicolumn{4}{c|}{91.0} &
				\multicolumn{4}{c|}{-} &
				\multicolumn{4}{c}{-} \\ \hline
				\multirow{3}{*}{T} &
				\multicolumn{8}{l|}{TMT\textreferencemark~\cite{liu2021video}} &
				\multicolumn{4}{c|}{85.8} &
				\multicolumn{4}{c|}{91.2} &
				\multicolumn{4}{c|}{-} &
				\multicolumn{4}{c}{-} \\
				&
				\multicolumn{8}{l|}{STT\textreferencemark~\cite{zhang2021spatiotemporal}} &
				\multicolumn{4}{c|}{86.3} &
				\multicolumn{4}{c|}{88.7} &
				\multicolumn{4}{c|}{\underline{97.4}} &
				\multicolumn{4}{c}{97.6} \\
				&
				\multicolumn{8}{l|}{DIL\textreferencemark~\cite{he2021dense}} &
				\multicolumn{4}{c|}{\underline{87.0}} &
				\multicolumn{4}{c|}{90.8} &
				\multicolumn{4}{c|}{97.1} &
				\multicolumn{4}{c}{97.6} \\ \cline{2-25}
				\multicolumn{1}{l|}{} &
				\multicolumn{8}{l|}{DCCT~(\textbf{Ours})} &
				\multicolumn{4}{c|}{\textbf{87.5}} &
				\multicolumn{4}{c|}{\textbf{92.3}} &
				\multicolumn{4}{c|}{\textbf{97.6}} &
				\multicolumn{4}{c}{\textbf{98.4}} \\ \hline
			\end{tabular}%
		}
		\vspace{-6mm}
	\end{center}
\end{table}
%-------------------------------------------------------------------------

\subsection{Comparison with State-of-the-arts}
In this subsection, we compare our approach with state-of-the-art methods on four video-based person Re-ID benchmarks.
Experimental results are reported in Tab.~\ref{table:mars duke sota} and Tab.~\ref{table:soa method}.
On the large-scale MARS and DukeMCMT-VID datasets, our method achieves the best results of $\textbf{87.5\%}$, $\textbf{97.6\%}$ in mAP and $\textbf{92.3\%}$, $\textbf{98.4\%}$ in Rank-1 accuracy, respectively.
Besides, on the iLIDS-VID and PRID2011 datasets, our method attains $\textbf{91.7\%}$ and $\textbf{96.8\%}$ Rank-1 accuracy, respectively.
They are much better than most state-of-the-art methods.
When compared with these methods, one can see that MGRA~\cite{zhang2020multi} and GRL~\cite{liu2021watching} also utilize the global representation to guide local feature refinement.
However, different from them, our method unifies CNNs and Transformers in a coupled structure and takes their global feature for mutual guidance for spatial complementary learning.
Thereby, our method surpasses MGRA and GRL by $\textbf{1.6\%}$ and $\textbf{2.7\%}$ in terms of mAP on MARS dataset, and by $\textbf{3.1\%}$ and $\textbf{0.3\%}$ in terms of Rank-1 accuracy on iLIDS-VID dataset.
Meanwhile, with pyramid spatial-temporal learning, PSTA~\cite{liu2021spatial} gains remarkable $91.5\%$ and $98.3\%$ Rank-1 accuracy on MARS and DukeMCMT-VID datasets.
Different from PSTA, we propose a novel spatial-temporal complementary learning framework to obtain richer video representations.
Thus, we obtain better performance than PSTA on MARS and DukeMCMT-VID, which validates the effectiveness of our proposed method.

When compared with some methods that utilize Transformer, our method also shows great superiority in performance.
For example, compared with STT~\cite{zhang2021spatiotemporal}, our method has $1.2\%$ and $3.6\%$ gains in mAP and Rank-1 accuracy.
Compared with recent DIL~\cite{he2021dense}, our method still achieves better results on MARS and DukeMCMT-VID.
Besides, TMT~\cite{liu2021video} introduces a multi-view Transformer to extract comprehensive video representations.
However, our method still performs better than TMT.
We note that all these methods add Transformer layers after CNN backbones to obtain enhanced representations.
Different from them, we deeply couple CNNs and Transformers, and propose a complementary content attention for spatial complementary learning.
Besides, the hierarchical temporal aggregation is leveraged to progressively integrate temporal information.
In this way, we utilize two kinds of typical visual features from the same videos and achieve spatial-temporal complementary learning for more informative representations.
Thereby, compared with previous methods, our method shows better performance on four video-based person Re-ID benchmarks.

%---------------------------------------------------------------
\begin{table}
	\begin{center}
		\doublerulesep=0.5pt
		\caption{Performance (\%) comparison on iLIDS-VID~\cite{wang2014person} and PRID-2011~\cite{hirzer2011person} datasets.
		}
		\label{table:soa method}
		%\vspace{-2mm}
		\resizebox{0.45\textwidth}{!}
		{
			\begin{tabular}{c|llllllll|clllclll|clllclll}
				\hline
				\multicolumn{1}{l|}{\multirow{2}{*}{}} &
				\multicolumn{8}{c|}{} &
				\multicolumn{8}{c|}{iLIDS-VID} &
				\multicolumn{8}{c}{PRID2011} \\
				\multicolumn{1}{l|}{} &
				\multicolumn{8}{c|}{Methods} &
				\multicolumn{4}{c|}{Rank-1} &
				\multicolumn{4}{c|}{Rank-5} &
				\multicolumn{4}{c|}{Rank-1} &
				\multicolumn{4}{c}{Rank-5} \\ \hline
				\multirow{14}{*}{C} &
				\multicolumn{8}{l|}{SeeForest~\cite{zhou2017see}} &
				\multicolumn{4}{c|}{55.2} &
				\multicolumn{4}{c|}{86.5} &
				\multicolumn{4}{c|}{79.4} &
				\multicolumn{4}{c}{94.4} \\
				&
				\multicolumn{8}{l|}{ASTPN~\cite{xu2017jointly}} &
				\multicolumn{4}{c|}{62.0} &
				\multicolumn{4}{c|}{86.0} &
				\multicolumn{4}{c|}{77.0} &
				\multicolumn{4}{c}{95.0} \\
				&
				\multicolumn{8}{l|}{Snippet~\cite{chen2018video}} &
				\multicolumn{4}{c|}{85.4} &
				\multicolumn{4}{c|}{96.7} &
				\multicolumn{4}{c|}{93.0} &
				\multicolumn{4}{c}{99.3} \\
				&
				\multicolumn{8}{l|}{STAN~\cite{li2018diversity}} &
				\multicolumn{4}{c|}{80.2} &
				\multicolumn{4}{c|}{-} &
				\multicolumn{4}{c|}{93.2} &
				\multicolumn{4}{c}{-} \\
				&
				\multicolumn{8}{l|}{STMP~\cite{liu2019spatial}} &
				\multicolumn{4}{c|}{84.3} &
				\multicolumn{4}{c|}{96.8} &
				\multicolumn{4}{c|}{92.7} &
				\multicolumn{4}{c}{98.8} \\
				&
				\multicolumn{8}{l|}{M3D~\cite{li2019multi}} &
				\multicolumn{4}{c|}{74.0} &
				\multicolumn{4}{c|}{94.3} &
				\multicolumn{4}{c|}{94.4} &
				\multicolumn{4}{c}{\textbf{100}} \\
				&
				\multicolumn{8}{l|}{Attribute~\cite{zhao2019attribute}} &
				\multicolumn{4}{c|}{86.3} &
				\multicolumn{4}{c|}{87.4} &
				\multicolumn{4}{c|}{93.9} &
				\multicolumn{4}{c}{99.5} \\
				&
				\multicolumn{8}{l|}{GLTR~\cite{li2019global}} &
				\multicolumn{4}{c|}{86.0} &
				\multicolumn{4}{c|}{98.0} &
				\multicolumn{4}{c|}{95.5} &
				\multicolumn{4}{c}{\textbf{100}} \\
				&
				\multicolumn{8}{l|}{FGRA~\cite{chen2020frame}} &
				\multicolumn{4}{c|}{88.0} &
				\multicolumn{4}{c|}{96.7} &
				\multicolumn{4}{c|}{95.5} &
				\multicolumn{4}{c}{\textbf{100}} \\
				&
				\multicolumn{8}{l|}{AMEM~\cite{li2020appearance}} &
				\multicolumn{4}{c|}{87.2} &
				\multicolumn{4}{c|}{97.7} &
				\multicolumn{4}{c|}{93.3} &
				\multicolumn{4}{c}{98.7} \\
				&
				\multicolumn{8}{l|}{MGRA~\cite{zhang2020multi}} &
				\multicolumn{4}{c|}{88.6} &
				\multicolumn{4}{c|}{98.0} &
				\multicolumn{4}{c|}{95.9} &
				\multicolumn{4}{c}{\underline{99.7}} \\
				&
				\multicolumn{8}{l|}{CTL~\cite{liu2021spatial}} &
				\multicolumn{4}{c|}{89.7} &
				\multicolumn{4}{c|}{97.0} &
				\multicolumn{4}{c|}{-} &
				\multicolumn{4}{c}{-} \\
				&
				\multicolumn{8}{l|}{GRL~\cite{liu2021watching}} &
				\multicolumn{4}{c|}{90.4} &
				\multicolumn{4}{c|}{\underline{98.3}} &
				\multicolumn{4}{c|}{96.2} &
				\multicolumn{4}{c}{\underline{99.7}} \\
				&
				\multicolumn{8}{l|}{PSTA~\cite{wang2021pyramid}} &
				\multicolumn{4}{c|}{91.5} &
				\multicolumn{4}{c|}{98.1} &
				\multicolumn{4}{c|}{95.6} &
				\multicolumn{4}{c}{98.9} \\
				&
				\multicolumn{8}{l|}{SINet~\cite{bai2022salient}} &
				\multicolumn{4}{c|}{\textbf{92.5}} &
				\multicolumn{4}{c|}{-} &
				\multicolumn{4}{c|}{96.5} &
				\multicolumn{4}{c}{-} \\ \hline
				\multirow{3}{*}{T} &
				\multicolumn{8}{l|}{TMT\textreferencemark~\cite{liu2021video}} &
				\multicolumn{4}{c|}{91.3} &
				\multicolumn{4}{c|}{\textbf{98.6}} &
				\multicolumn{4}{c|}{96.4} &
				\multicolumn{4}{c}{99.3} \\
				&
				\multicolumn{8}{l|}{DIL\textreferencemark~\cite{he2021dense}} &
				\multicolumn{4}{c|}{\underline{92.0}} &
				\multicolumn{4}{c|}{96.0} &
				\multicolumn{4}{c|}{-} &
				\multicolumn{4}{c}{-} \\ \cline{2-25}
				&
				\multicolumn{8}{l|}{DCCT~(\textbf{Ours})} &
				\multicolumn{4}{c|}{91.7} &
				\multicolumn{4}{c|}{\textbf{98.6}} &
				\multicolumn{4}{c|}{\textbf{96.8}} &
				\multicolumn{4}{c}{\underline{99.7}} \\ \hline
			\end{tabular}%
		}
		\vspace{-6mm}
	\end{center}
\end{table}
%---------------------------------------------------------------------

\subsection{Ablation Study}
In this subsection, we conduct ablation experiments mainly on MARS to investigate the effectiveness of our method.

%----------------------------------------------------------------------------
\begin{table}[]
	\begin{center}
		\caption{Ablation results with different backbone networks.}
		%\vspace{-2mm}
		\label{tab:ablation complementarity}
		\resizebox{0.46\textwidth}{!}{%
			\begin{tabular}{cc|cccc}
				\hline
				\multicolumn{2}{c|}{Backbone}                   & \multicolumn{4}{c}{MARS}         \\ \hline
				\multicolumn{1}{c|}{CNN}          & Transformer & mAP  & Rank-1 & Rank-5 & Rank-20 \\ \hline
				\multicolumn{1}{c|}{MobileNet-v2} & --          & 80.1 & 88.0   & 95.5   & 97.9    \\
				\multicolumn{1}{c|}{--}           & DeiT-Tiny   & 75.7 & 85.6   & 94.2   & 97.1    \\ \hline
				\multicolumn{2}{c|}{Coupled}                 & 82.3 & 89.7   & 96.4   & 98.3    \\ \hline
				\multicolumn{1}{c|}{ResNet-50}    & --          & 82.4 & 89.9   & 96.3   & 98.2    \\
				\multicolumn{1}{c|}{--}           & ViT-Base    & 83.6 & 90.4   & 96.4   & 98.5    \\ \hline
				\multicolumn{2}{c|}{Coupled}                 & 85.0 & 91.1   & 97.0   & 98.6    \\ \hline
				\multicolumn{1}{c|}{DenseNet-121}    & --          & 83.0 & 90.5   & 96.0   & 98.3    \\
				\multicolumn{1}{c|}{--}           & Swin-Large  & 83.3 & 89.8   & 96.1   & 98.6    \\ \hline
				\multicolumn{2}{c|}{Coupled}                 & 86.3 & 91.5   & 97.4   & 98.9    \\ \hline
			\end{tabular}%
		}
	\end{center}
	\vspace{-2mm}
\end{table}
%---------------------------------------------------------------
%---------------------------------------------------------------

\begin{table}[]
	\begin{center}
		\caption{Ablation results of key components on MARS.}
		\label{tab:ablation key components}
		\resizebox{0.45\textwidth}{!}{%
			\begin{tabular}{l|cc|cc}
				\hline
				Methods   & mAP  & Rank-1 & FLOPs (G) & Params (M) \\ \hline
				ResNet-50 & 82.4 & 89.9   & 3.26      & 22.42      \\
				ViT-Base  & 83.6 & 90.4   & 8.83      & 81.64      \\ \hline
				Coupled   & 85.0 & 91.1   & 12.09     & 104.07     \\
				+ CCA     & 85.5 & 91.4   & 13.07     & 125.54     \\
				+ HTA     & 87.2 & 92.1   & 13.15     & 222.61     \\
				+ KD      & 87.5 & 92.3   & 13.15     & 222.61     \\ \hline
			\end{tabular}%
		}
	\end{center}
	\vspace{-6mm}
\end{table}

\textbf{Complementarity between CNNs and Transformers.}
In our proposed method, we take advantages of CNNs and Transformers for complementary learning.
To verify the complementarity, we adopt different structure settings for this ablation.
More specifically, we utilize the MobileNet-V2~\cite{sandler2018mobilenetv2}, ResNet-50~\cite{he2016deep} and DenseNet-121~\cite{huang2017densely} to obtain CNN-based representations.
Correspondingly, we employ DeiT-Tiny~\cite{touvron2021training}, ViT-Base~\cite{dosovitskiy2020image} and Swin-Large~\cite{liu2021swin} to obtain Transformer-based representations.
For these networks, we drop their fully-connected layers and apply GAP and TAP to obtain the final video feature vector.
In the coupled structure, the video feature vectors from CNNs and Transformers are concatenated for training and test.
The validation results are shown in Tab.~\ref{tab:ablation complementarity}.
From the experimental results, we can see that, compared with the independent representation, the performances are significantly improved when unifying CNNs and Transformers.
For example, when coupling ResNet-50 and ViT-Base, the performance could attain $85.0\%$ and $91.1\%$ in terms of mAP and Rank-1 accuracy on MARS, respectively.
Meanwhile, Fig.~\ref{fig:visualization_complementarity} displays their visual differences of feature maps in ResNet-50 and ViT-Base.
The results validate the distinctiveness and complementarity between CNN-based features and Transformer-based features.
If not specified, we use the coupled ResNet-50 and ViT-Base as our baseline in the remaining experiments.

\textbf{Effects of Key Components.}
We gradually add the proposed key components to the baseline for ablation analysis.
The ablation results are reported in Tab.~\ref{tab:ablation key components}.
Here, ``Coupled'' represents the baseline of unifying ResNet-50 and ViT-Base.
``+ CCA'' indicates the proposed CCA is utilized.
The mAP and Rank-1 accuracy are improved by $0.5\%$ and $0.3\%$ on MARS, respectively.
The improvements illustrate that the mutual guidance between CNNs and Transformers with our CCA is beneficial for complementary learning.
Our CCA helps to get richer spatial representations.
``+ HTA'' means two-layers HTA are deployed after CCA.
Compared with ``+ CCA'', both mAP and Rank-1 accuracy are significantly improved.
These improvements indicate that our spatial-temporal complementary learning after coupled Convolution-Transformer
is beneficial to extract better video representations.
Meanwhile, we deploy self-distillation training scheme in ``+ KD'', leading to further improvements.
Eventually, compared with ``Coupled'', the whole framework attains higher performance.

%-------------------------------------------------------------
\begin{table}[]
	\begin{center}
		\caption{Ablation results of the proposed CCA on MARS.}
		%\vspace{-2mm}
		\label{tab:ablation_CCA}
		\resizebox{0.44\textwidth}{!}{%
			\begin{tabular}{l|cc|cc}
				\hline
				& \multicolumn{2}{c|}{ResNet-50} & \multicolumn{2}{c}{ViT-Base} \\ \hline
				\multicolumn{1}{c|}{Methods} & mAP           & Rank-1         & mAP          & Rank-1        \\ \hline
				Backbone                     & 82.4          & 89.9           & 83.6         & 90.4          \\ \hline
				Backbone + CCA               & 83.6          & 90.8           & 84.3         & 91.2          \\
				- SH                    & 83.2          & 90.5           & 84.1         & 91.0          \\
				- CH                    & 82.7          & 90.3           & 83.8         & 90.7          \\ \hline
			\end{tabular}%
		}
	\end{center}
	\vspace{-2mm}
\end{table}
%%----------------------------------------------------------
%---------------------------------------------------------------
\begin{table}[]
	\begin{center}
		\caption{Ablation results of the Proposed HTA on MARS.}
		\label{tab:ablation_HTA}
		\resizebox{0.45\textwidth}{!}{%
			\begin{tabular}{l|ccc|cc}
				\hline
				Methods                          & TT           & AT           & GA           & mAP  & Rank-1 \\ \hline
				Coupled + CCA                    & $\times$     & $\times$     & $\times$     & 85.5 & 91.4   \\ \hline
				\multirow{3}{*}{+ HTA (one depth)} & $\checkmark$ & $\times$     & $\times$     & 85.9 & 91.5   \\ \cline{5-6}
				& $\checkmark$ & $\checkmark$ & $\times$     & 86.5 & 91.6   \\ \cline{5-6}
				& $\checkmark$ & $\checkmark$ & $\checkmark$ & 86.8 & 91.8   \\ \hline
			\end{tabular}%
		}
	\end{center}
	\vspace{-6mm}
\end{table}
%---------------------------------------------------------------

\textbf{Analysis of Computational Cost.}
The ablation analysis of computational complexity and memory is also reported in Tab.~\ref{tab:ablation key components}.
With increased accuracies, the computational cost is raised to somewhat higher.
Noted that, our FLOPs are not much high. However, due to the high dimension of output features in ResNet-50 and ViT-Base, the parameters increase more.
In the future, more techniques will be explored to address this problem, such as reducing the information redundancy of features in channels.

\textbf{Effects of SH and CH in CCA.}
We separately investigate the effectiveness of SH and CH in ResNet-50 and ViT-Base.
The results on MARS are shown in Tab.~\ref{tab:ablation_CCA}.
``- SH'' or ``- CH'' means that the self-attention head or the cross-attention head is removed from CCA, respectively.
In the ResNet-50 branch, compared with ``Backbone + CCA'', ``- SH'' and ``- CH'' show some accuracy degradations.
It suggests that under the guidance of global features, spatial features could capture more discriminative cues from the whole video.
In the ViT-Base branch, when removing the SH, the performance has a slight drop.
However, when removing the CH, the Rank-1 accuracy has a significant drop.
The reason may be that the features in Transformers have captured global observations.
Then, under the guidance of convolutional features, the CH will give more attention on meaningful regions and reduce the influence of noises for higher performance.
Moreover, the visualizations in Fig.~\ref{fig:visualization_CCA} help us to better understand these facts.

\textbf{Effects of TT, AT and GA in HTA.}
We also conduct experiments to explore the effects of different modules in HTA.
For experiments, we gradually employ TT, AT and GA after CCA to realize temporal embedding and complementary attention, respectively.
The ablation results are reported in Tab.~\ref{tab:ablation_HTA}.
Compared with ``Baseline + CCA'', TT and AT help to gain $1.0\%$ mAP and $0.4\%$ Rank-1 accuracy on MARS.
The improvements suggest that inter-frame relations are useful to encode temporal information.
Besides, our GA helps to communicate the aggregated video feature to two branches for temporal complementary.
Therefore, it helps to get higher similarity scores, leading to a better mAP.
%

%------------------------------------------------
\begin{figure}[t]
	\centering
	\resizebox{0.4\textwidth}{!}
	{
		\begin{tabular}{@{}c@{}c@{}}
			\includegraphics[width=0.5\linewidth,height=0.4\linewidth]{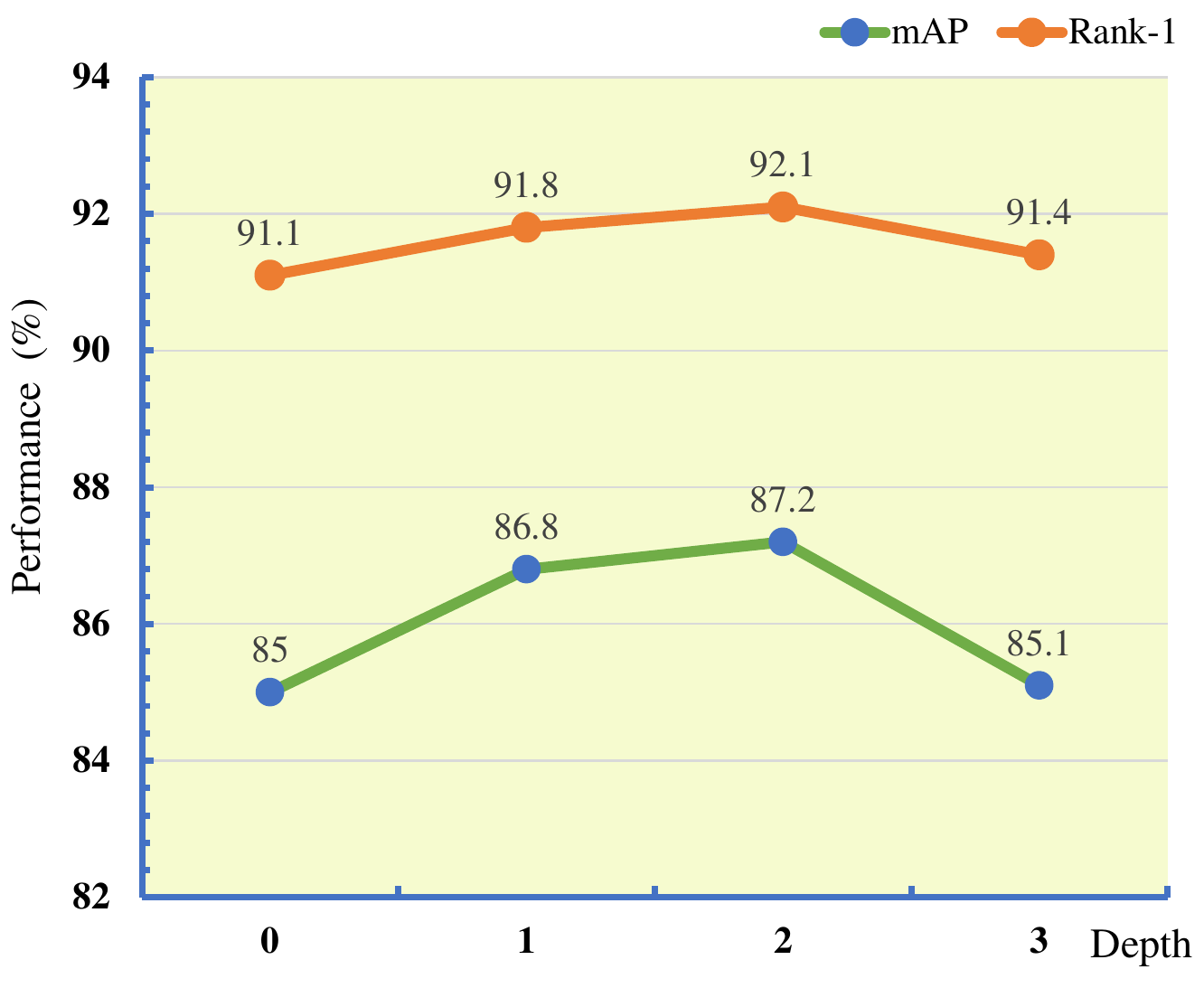} \\
		\end{tabular}
	}
	\caption{Ablation results with different depths of HTA on MARS.
	}
	\label{fig:Depth}
	%\vspace{-2mm}
\end{figure}
%------------------------------------------------------------

%-------------------------------------------------------------------------
% Please add the following required packages to your document preamble:
% \usepackage{graphicx}
\begin{table}[t]
	\caption{Ablation results with different aggregation methods on MARS.}
	\label{tab:aggregation_technologies}
	\vspace{-4mm}
	\begin{center}
		\resizebox{0.45\textwidth}{!}{%
			\begin{tabular}{c|ll|cc}
				\hline
				Backbone              & Spatial & Temporal & \multicolumn{1}{c}{mAP} & \multicolumn{1}{c}{Rank-1} \\ \hline
				\multicolumn{1}{l|}{} & + GAP   & + TAP    & 85.0                    & 91.1                       \\
				ResNet50              & + SAP   & + HTA    & 86.3                    & 91.6                       \\
				\& ViT-Base           & + CCA   & + LSTM   & 85.8                    & 91.4                       \\
				& + CCA   & + HTA    & 87.2                    & 92.1                       \\ \hline
			\end{tabular}%
		}
		\vspace{-6mm}
	\end{center}
\end{table}
%--------------------------------------------------------------

\textbf{Effects of Different Depths in HTA.}
The results in Fig.~\ref{fig:Depth} show the effects of different depths with HTA on MARS.
When the depth is 0, the method is equivalent to ``+ CCA'' in Tab.~\ref{tab:ablation key components}.
From the results, we can see that increasing the depth of HTA gains better performance and the depth of 2 gets the best performance.
When the depth is 2, the gated attention module is utilized to adaptively deliver the aggregated features for complementary learning.
It brings clear improvements in terms of mAP.
However, when the depth is more than 2, the performance has some degradations.
The reason is that a deeper HTA may lead to overfitting.
In order to balance the accuracy and efficiency, we set the depth of HTA to be 2 by default if not specifically stated in experiments.

\textbf{Effects of Different Aggregation Methods.}
To verify the superior effectiveness of our CCA and HTA, we utilize other spatial-temporal feature aggregation methods to replace our CCA and HTA for comparison.
The results are shown in Tab.~\ref{tab:aggregation_technologies}.
In the 2nd row of this table, GAP and TAP are applied for direct spatial-temporal feature aggregation.
In the 5th row of this table, our CCA and HTA are utilized and attain significant improvements.
However, when replacing our CCA with Spatial Attention Pooling (SAP)~\cite{liu2021video} or replacing our HTA with LSTM~\cite{dai2019video}, the mAP has some degradations.
The results are shown in the 3rd and 4th rows of Tab.~\ref{tab:aggregation_technologies}.
These results demonstrate that our spatial-temporal complementary leaning with CCA and HTA helps to fully take their advantages for enhanced video representations.

\begin{figure}
	\centering
	\resizebox{0.48\textwidth}{!}
	{
		\begin{tabular}{@{}c@{}c@{}}
			\includegraphics[width=0.5\linewidth,height=0.33\linewidth]{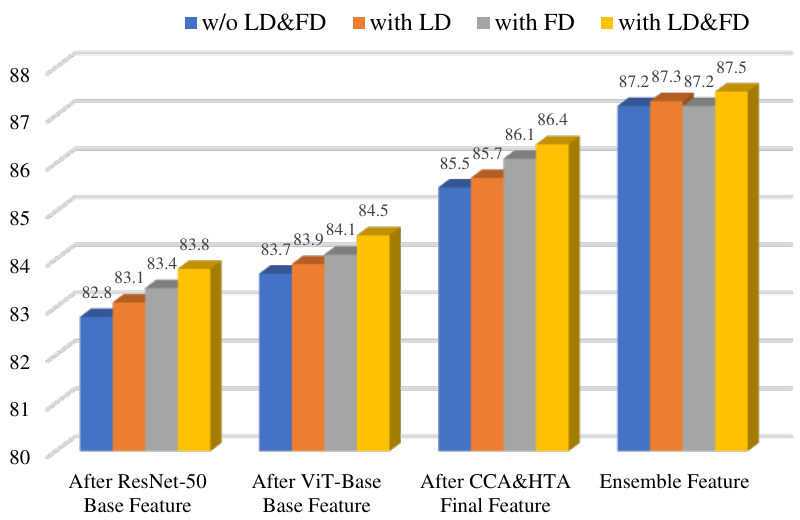} \\
		\end{tabular}
	}
	%\vspace{-2mm}
	\caption{Ablation results of self-distillation on MARS.
	}
	\label{fig:SelfDistillation}
	\vspace{-4mm}
\end{figure}

\textbf{Effects of Self-Distillation Training Strategy.}
In this work, we introduce a self-distillation learning scheme for model training, which contains Logistic Distillation (LD) and Feature Distillation (FD).
The ablation results are shown in Fig.~\ref{fig:SelfDistillation}.
Firstly, we can find that the ensemble of base features and final feature can obtain remarkable performances with two-stage supervision.
Besides, the performance can be further improved by our self-distillation training strategy.
Both LD and FD are beneficial for ResNet-50 and ViT-Base to attain higher accuracy.
In addition, compared with LD, FD has a greater impact on discriminative representations.
The accuracy can be further improved with the joint utilization of the LD and FD.
It also indicates that the aggregated features after CCA and HTA, as a teacher, can guide the CNN or Transformer backbones to obtain better representations.
Meanwhile, after self-distillation training, if removing our CCA and HTA, the base features also attain better performance than baseline.
It indicates our self-distillation training strategy could help to reduce the inference cost for higher efficiency.
For other video-based visual tasks, the proposed self-distillation strategy can be easily adopted and is beneficial to train deep and complex spatial-temporal learning networks.

%------------------------------------------------
%------------------------------------------------------------

\subsection{Visualization Analysis}
In this subsection, the feature maps and attention maps from our network are visualized to explain the effectiveness of our approach.
First, we compare the feature maps from CNNs and Transformers to verify their complementary.
Besides, we visualize the attentions maps in our CCA to illustrate the spatial attention and complementary learning.

\begin{figure}[t]
	\centering
	\resizebox{0.48\textwidth}{!}
	{
		\begin{tabular}{@{}c@{}c@{}}
			\includegraphics[width=1.0\linewidth,height=0.6\linewidth]{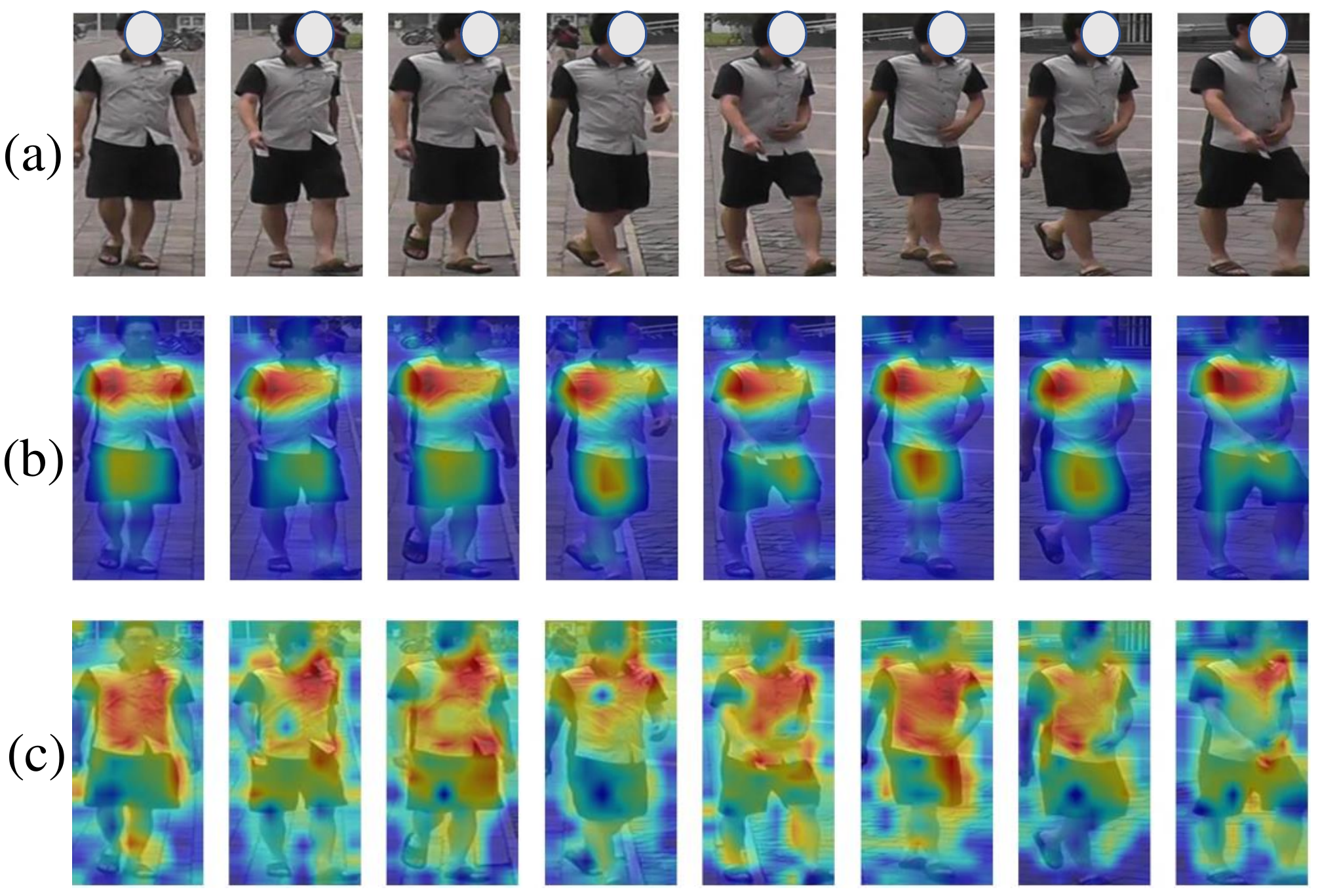} \\
		\end{tabular}
	}
	%\vspace{-2mm}
	\caption{Comparison of the final features maps from ResNet-50 and ViT-Base with Grad-CAM~\cite{selvaraju2017grad}. (a) Raw images in a video; (b) Feature maps of ResNet-50; (c) Feature maps of ViT-Base. The highlighted regions in red present the discriminative regions.
	}
	\label{fig:visualization_complementarity}
	\vspace{-4mm}
\end{figure}

\textbf{Discriminative Feature Maps in CNNs and Transformers.}
In Fig.~\ref{fig:visualization_complementarity}, we visualize the heat maps of final features from ResNet-50 and ViT-Base with Grad-CAM~\cite{selvaraju2017grad}.
Comparing two kinds of typical features from the same sequences, we can find that, their highlighted areas are very different.
For example, the high lightness in ResNet-50 tends to focus on the main parts of the human body.
While ViT-Base always attempts to capture more visual cues with the global perception.
The visual differences have verified the complementarity between ResNet-50 and ViT-Base.
Based on the complementarity between CNNs and Transformers, in this work, we propose a novel complementary learning framework to take fully their different abilities for better video representations.

%%----------------------------------------------------------
\begin{figure}[t]
	\centering
	\resizebox{0.48\textwidth}{!}
	{
		\begin{tabular}{@{}c@{}c@{}}
			\includegraphics[width=1.0\linewidth,height=0.9\linewidth]{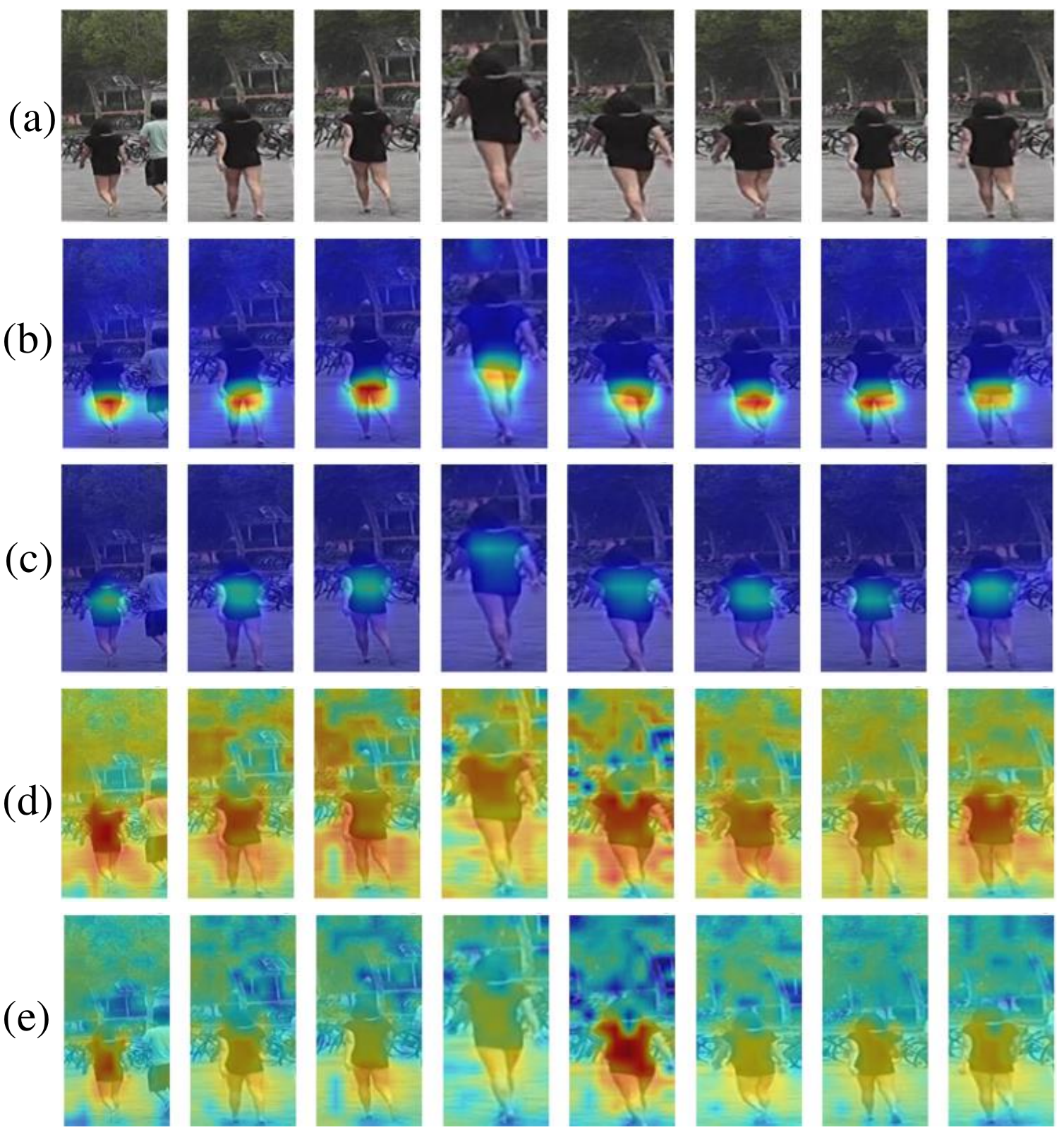} \\
		\end{tabular}
	}
	%\vspace{-2mm}
	\caption{The attention maps from SH and CH in our proposed CCA. (a) Raw images in a video; (b) Self-attention maps in ResNet-50; (c) Cross-attention maps in ResNet-50; (d) Self-attention maps in ViT-Base; (e) Cross-attention maps in ViT-Base. The deeper red means bigger attention weights.
	}
	\label{fig:visualization_CCA}
	%\vspace{-4mm}
\end{figure}
%--------------------------------------------------

%----------------------------------
%------------------------------------------------

\textbf{Visualization of Attention Maps in CCA.}
We also show the visualizations to verify the effects of the proposed CCA module.
Specifically, the self-attention maps in SH are shown in Fig.~\ref{fig:visualization_CCA}(a) and Fig.~\ref{fig:visualization_CCA}(c).
We can find that the SH helps CNNs to focus on the most discriminative regions and helps Transformers to capture the global content of the whole image.
In this way, the spatial features are enhanced with global guidance.
The cross-attention maps in CH are shown in Fig.~\ref{fig:visualization_CCA}(b) and Fig.~\ref{fig:visualization_CCA}(d).
We can see that, Transformer-based representations guide CNNs to capture more different yet meaningful visual information.
Meanwhile, CNN-based representations help Transformers to pay more attention on the main parts of pedestrians and reduce the impact of noises in the background.
Thus, CH could achieve complementary learning by the mutual guidance of CNNs and Transformers.
In this way, our proposed CCA achieves the feature enhancement and captures the complementary information from CNNs and Transformers for richer feature representations.
\section{Conclusion and Future Works}
In this work, we propose a spatial-temporal complementary learning framework named Deeply-Coupled Convolution-Transformer (DCCT) for video-based person Re-ID.
Motivated by the differences between CNNs and Transformers, we couple them to extract two kinds of visual features and experimentally verify their complementarity.
Then, we propose a Complementary Content Attention (CCA) to enhance spatial feature representations, and a Hierarchical Temporal Aggregation (HTA) to integrate temporal information.
Both of them are very effective in extracting robust representations by complementary learning.
Besides, a self-distillation training strategy is utilized to optimize the whole framework for higher accuracy and more efficiency.
Extensive experiments on four public Re-ID benchmarks demonstrate that our framework performs better than most state-of-the-art methods.
In the future, more techniques will be explored to reduce the computational cost.
Besides, how to unify CNNs and Transformers more efficiently deserves further research, such as the mutual learning.

%%%------------------------------------------------------------------
\ifCLASSOPTIONcaptionsoff
  \newpage
\fi
\bibliographystyle{IEEEtran}
\bibliography{refs}

% Generated by IEEEtran.bst, version: 1.14 (2015/08/26)
\begin{thebibliography}{10}
\providecommand{\url}[1]{#1}
\csname url@samestyle\endcsname
\providecommand{\newblock}{\relax}
\providecommand{\bibinfo}[2]{#2}
\providecommand{\BIBentrySTDinterwordspacing}{\spaceskip=0pt\relax}
\providecommand{\BIBentryALTinterwordstretchfactor}{4}
\providecommand{\BIBentryALTinterwordspacing}{\spaceskip=\fontdimen2\font plus
\BIBentryALTinterwordstretchfactor\fontdimen3\font minus
  \fontdimen4\font\relax}
\providecommand{\BIBforeignlanguage}[2]{{%
\expandafter\ifx\csname l@#1\endcsname\relax
\typeout{** WARNING: IEEEtran.bst: No hyphenation pattern has been}%
\typeout{** loaded for the language `#1'. Using the pattern for}%
\typeout{** the default language instead.}%
\else
\language=\csname l@#1\endcsname
\fi
#2}}
\providecommand{\BIBdecl}{\relax}
\BIBdecl

\bibitem{he2016deep}
K.~He, X.~Zhang, S.~Ren, and J.~Sun, ``Deep residual learning for image
  recognition,'' in \emph{Proceedings of the IEEE Conference on Computer Vision
  and Pattern Recognition}, 2016, pp. 770--778.

\bibitem{dosovitskiy2020image}
A.~Dosovitskiy, L.~Beyer, A.~Kolesnikov, D.~Weissenborn, X.~Zhai,
  T.~Unterthiner, M.~Dehghani, M.~Minderer, G.~Heigold, S.~Gelly \emph{et~al.},
  ``An image is worth 16x16 words: Transformers for image recognition at
  scale,'' in \emph{International Conference on Learning Representations},
  2020, pp. 1--13.

\bibitem{cheng2016person}
D.~Cheng, Y.~Gong, S.~Zhou, J.~Wang, and N.~Zheng, ``Person re-identification
  by multi-channel parts-based cnn with improved triplet loss function,'' in
  \emph{Proceedings of the IEEE Conference on Computer Vision and Pattern
  Recognition}, 2016, pp. 1335--1344.

\bibitem{wang2021pyramid}
Y.~Wang, P.~Zhang, S.~Gao, X.~Geng, H.~Lu, and D.~Wang, ``Pyramid
  spatial-temporal aggregation for video-based person re-identification,'' in
  \emph{Proceedings of the IEEE Conference on Computer Vision and Pattern
  Recognition}, 2021, pp. 12\,026--12\,035.

\bibitem{ye2021dynamic}
M.~Ye, C.~Chen, J.~Shen, and L.~Shao, ``Dynamic tri-level relation mining with
  attentive graph for visible infrared re-identification,'' \emph{IEEE
  Transactions on Information Forensics and Security}, vol.~17, pp. 386--398,
  2021.

\bibitem{wu2021person}
D.~Wu, M.~Ye, G.~Lin, X.~Gao, and J.~Shen, ``Person re-identification by
  context-aware part attention and multi-head collaborative learning,''
  \emph{IEEE Transactions on Information Forensics and Security}, vol.~17, pp.
  115--126, 2021.

\bibitem{wang2022Pro}
Y.~Wang, J.~Peng, H.~Wang, and M.~Wang, ``Progressive learning with multi-scale
  attention network for cross-domain vehicle re-identification,'' \emph{Science
  China Information Sciences}, vol.~65, p. 160103, 2022.

\bibitem{liu2021video}
X.~Liu, P.~Zhang, C.~Yu, H.~Lu, X.~Qian, and X.~Yang, ``A video is worth three
  views: Trigeminal transformers for video-based person re-identification,''
  \emph{arXiv:2104.01745}, 2021.

\bibitem{xiao2021early}
T.~Xiao, P.~Dollar, M.~Singh, E.~Mintun, T.~Darrell, and R.~Girshick, ``Early
  convolutions help transformers see better,'' \emph{Advances in Neural
  Information Processing Systems}, vol.~34, 2021.

\bibitem{d2021convit}
S.~d’Ascoli, H.~Touvron, M.~L. Leavitt, A.~S. Morcos, G.~Biroli, and
  L.~Sagun, ``Convit: Improving vision transformers with soft convolutional
  inductive biases,'' in \emph{International Conference on Machine
  Learning}.\hskip 1em plus 0.5em minus 0.4em\relax PMLR, 2021, pp. 2286--2296.

\bibitem{zhang2021spatiotemporal}
T.~Zhang, L.~Wei, L.~Xie, Z.~Zhuang, Y.~Zhang, B.~Li, and Q.~Tian,
  ``Spatiotemporal transformer for video-based person re-identification,''
  \emph{arXiv:2103.16469}, 2021.

\bibitem{he2021dense}
T.~He, X.~Jin, X.~Shen, J.~Huang, Z.~Chen, and X.-S. Hua, ``Dense interaction
  learning for video-based person re-identification,'' in \emph{Proceedings of
  the IEEE International Conference on Computer Vision}, 2021, pp. 1490--1501.

\bibitem{peng2021conformer}
Z.~Peng, W.~Huang, S.~Gu, L.~Xie, Y.~Wang, J.~Jiao, and Q.~Ye, ``Conformer:
  Local features coupling global representations for visual recognition,'' in
  \emph{Proceedings of the IEEE International Conference on Computer Vision},
  2021, pp. 367--376.

\bibitem{li2018deep}
Z.~Li, J.~Tang, and T.~Mei, ``Deep collaborative embedding for social image
  understanding,'' \emph{IEEE Transactions on Pattern Analysis and Machine
  Intelligence}, vol.~41, pp. 2070--2083, 2018.

\bibitem{li2021ctnet}
Z.~Li, Y.~Sun, L.~Zhang, and J.~Tang, ``Ctnet: Context-based tandem network for
  semantic segmentation,'' \emph{IEEE Transactions on Pattern Analysis and
  Machine Intelligence}, vol.~44, pp. 9904--9917, 2021.

\bibitem{jin2020deep}
L.~Jin, Z.~Li, and J.~Tang, ``Deep semantic multimodal hashing network for
  scalable image-text and video-text retrievals,'' \emph{IEEE Transactions on
  Neural Networks and Learning Systems}, 2020.

\bibitem{li2020weakly}
Z.~Li, J.~Tang, L.~Zhang, and J.~Yang, ``Weakly-supervised semantic guided
  hashing for social image retrieval,'' \emph{International Journal of Computer
  Vision}, vol. 128, pp. 2265--2278, 2020.

\bibitem{ye2020augmentation}
M.~Ye, J.~Shen, X.~Zhang, P.~C. Yuen, and S.-F. Chang, ``Augmentation invariant
  and instance spreading feature for softmax embedding,'' \emph{IEEE
  Transactions on Pattern Analysis and Machine Intelligence}, vol.~44, pp.
  924--939, 2020.

\bibitem{fu2019sta}
Y.~Fu, X.~Wang, Y.~Wei, and T.~Huang, ``Sta: Spatial-temporal attention for
  large-scale video-based person re-identification,'' in \emph{Proceedings of
  the AAAI Conference on Artificial Intelligence}, 2019, pp. 8287--8294.

\bibitem{gu2020appearance}
X.~Gu, H.~Chang, B.~Ma, H.~Zhang, and X.~Chen, ``Appearance-preserving 3d
  convolution for video-based person re-identification,'' in \emph{Proceedings
  of the European Conference on Computer Vision}, 2020, pp. 228--243.

\bibitem{zhang2020multi}
Z.~Zhang, C.~Lan, W.~Zeng, and Z.~Chen, ``Multi-granularity reference-aided
  attentive feature aggregation for video-based person re-identification,'' in
  \emph{Proceedings of the IEEE Conference on Computer Vision and Pattern
  Recognition}, 2020, pp. 10\,407--10\,416.

\bibitem{chen2020temporal}
G.~Chen, Y.~Rao, J.~Lu, and J.~Zhou, ``Temporal coherence or temporal motion:
  Which is more critical for video-based person re-identification?'' in
  \emph{Proceedings of the European Conference on Computer Vision}, 2020, pp.
  660--676.

\bibitem{bai2022salient}
S.~Bai, B.~Ma, H.~Chang, R.~Huang, and X.~Chen, ``Salient-to-broad transition
  for video person re-identification,'' in \emph{Proceedings of the IEEE
  Conference on Computer Vision and Pattern Recognition}, 2022, pp. 7339--7348.

\bibitem{liu2021watching}
X.~Liu, P.~Zhang, C.~Yu, H.~Lu, and X.~Yang, ``Watching you: Global-guided
  reciprocal learning for video-based person re-identification,'' in
  \emph{Proceedings of the IEEE Conference on Computer Vision and Pattern
  Recognition}, 2021, pp. 13\,334--13\,343.

\bibitem{li2019multi}
J.~Li, S.~Zhang, and T.~Huang, ``Multi-scale 3d convolution network for video
  based person re-identification,'' in \emph{Proceedings of the AAAI Conference
  on Artificial Intelligence}, 2019, pp. 8618--8625.

\bibitem{hou2020temporal}
R.~Hou, H.~Chang, B.~Ma, S.~Shan, and X.~Chen, ``Temporal complementary
  learning for video person re-identification,'' in \emph{Proceedings of the
  European Conference on Computer Vision}, 2020, pp. 388--405.

\bibitem{liu2021swin}
Z.~Liu, Y.~Lin, Y.~Cao, H.~Hu, Y.~Wei, Z.~Zhang, S.~Lin, and B.~Guo, ``Swin
  transformer: Hierarchical vision transformer using shifted windows,''
  \emph{arXiv:2103.14030}, 2021.

\bibitem{carion2020end}
N.~Carion, F.~Massa, G.~Synnaeve, N.~Usunier, A.~Kirillov, and S.~Zagoruyko,
  ``End-to-end object detection with transformers,'' in \emph{Proceedings of
  the European Conference on Computer Vision}, 2020, pp. 213--229.

\bibitem{wang2021end}
Y.~Wang, Z.~Xu, X.~Wang, C.~Shen, B.~Cheng, H.~Shen, and H.~Xia, ``End-to-end
  video instance segmentation with transformers,'' in \emph{Proceedings of the
  IEEE Conference on Computer Vision and Pattern Recognition}, 2021, pp.
  8741--8750.

\bibitem{zeng2020learning}
Y.~Zeng, J.~Fu, and H.~Chao, ``Learning joint spatial-temporal transformations
  for video inpainting,'' in \emph{Proceedings of the European Conference on
  Computer Vision}, 2020, pp. 528--543.

\bibitem{he2021transreid}
S.~He, H.~Luo, P.~Wang, F.~Wang, H.~Li, and W.~Jiang, ``Transreid:
  Transformer-based object re-identification,'' \emph{Proceedings of the IEEE
  International Conference on Computer Vision}, 2021.

\bibitem{zhang2021hat}
G.~Zhang, P.~Zhang, J.~Qi, and H.~Lu, ``Hat: Hierarchical aggregation
  transformers for person re-identification,'' in \emph{Proceedings of the 29th
  ACM International Conference on Multimedia}, 2021, pp. 516--525.

\bibitem{hinton2015distilling}
G.~Hinton, O.~Vinyals, and J.~Dean, ``Distilling the knowledge in a neural
  network,'' \emph{arXiv:1503.02531}, 2015.

\bibitem{zhang2019your}
L.~Zhang, J.~Song, A.~Gao, J.~Chen, C.~Bao, and K.~Ma, ``Be your own teacher:
  Improve the performance of convolutional neural networks via self
  distillation,'' in \emph{Proceedings of the IEEE International Conference on
  Computer Vision}, 2019, pp. 3713--3722.

\bibitem{romero2014fitnets}
A.~Romero, N.~Ballas, S.~E. Kahou, A.~Chassang, C.~Gatta, and Y.~Bengio,
  ``Fitnets: Hints for thin deep nets,'' \emph{arXiv:1412.6550}, 2014.

\bibitem{tung2019similarity}
F.~Tung and G.~Mori, ``Similarity-preserving knowledge distillation,'' in
  \emph{Proceedings of the IEEE International Conference on Computer Vision},
  2019, pp. 1365--1374.

\bibitem{park2019relational}
W.~Park, D.~Kim, Y.~Lu, and M.~Cho, ``Relational knowledge distillation,'' in
  \emph{Proceedings of the IEEE Conference on Computer Vision and Pattern
  Recognition}, 2019, pp. 3967--3976.

\bibitem{zheng2021pose}
K.~Zheng, C.~Lan, W.~Zeng, J.~Liu, Z.~Zhang, and Z.-J. Zha, ``Pose-guided
  feature learning with knowledge distillation for occluded person
  re-identification,'' in \emph{Proceedings of the 29th ACM International
  Conference on Multimedia}, 2021, pp. 4537--4545.

\bibitem{zhang2018deep}
Y.~Zhang, T.~Xiang, T.~M. Hospedales, and H.~Lu, ``Deep mutual learning,'' in
  \emph{Proceedings of the IEEE Conference on Computer Vision and Pattern
  Recognition}, 2018, pp. 4320--4328.

\bibitem{ge2020mutual}
Y.~Ge, D.~Chen, and H.~Li, ``Mutual mean-teaching: Pseudo label refinery for
  unsupervised domain adaptation on person re-identification,''
  \emph{arXiv:2001.01526}, 2020.

\bibitem{zhou2019discriminative}
S.~Zhou, F.~Wang, Z.~Huang, and J.~Wang, ``Discriminative feature learning with
  consistent attention regularization for person re-identification,'' in
  \emph{Proceedings of the IEEE International Conference on Computer Vision},
  2019, pp. 8040--8049.

\bibitem{xiao2017joint}
T.~Xiao, S.~Li, B.~Wang, L.~Lin, and X.~Wang, ``Joint detection and
  identification feature learning for person search,'' in \emph{Proceedings of
  the IEEE Conference on Computer Vision and Pattern Recognition}, 2017, pp.
  3415--3424.

\bibitem{ding2015deep}
S.~Ding, L.~Lin, G.~Wang, and H.~Chao, ``Deep feature learning with relative
  distance comparison for person re-identification,'' \emph{Pattern
  Recognition}, vol.~48, no.~10, pp. 2993--3003, 2015.

\bibitem{kullback1951on}
S.~Kullback and R.~Leibler, ``On information and sufficiency,'' \emph{Annals of
  Mathematical Statistics}, pp. 79--86, 1951.

\bibitem{zheng2016mars}
L.~Zheng, Z.~Bie, Y.~Sun, J.~Wang, C.~Su, S.~Wang, and Q.~Tian, ``Mars: A video
  benchmark for large-scale person re-identification,'' in \emph{Proceedings of
  the European Conference on Computer Vision}, 2016, pp. 868--884.

\bibitem{zheng2017unlabeled}
Z.~Zheng, L.~Zheng, and Y.~Yang, ``Unlabeled samples generated by gan improve
  the person re-identification baseline in vitro,'' in \emph{Proceedings of the
  IEEE International Conference on Computer Vision}, 2017, pp. 3754--3762.

\bibitem{wang2014person}
T.~Wang, S.~Gong, X.~Zhu, and S.~Wang, ``Person re-identification by video
  ranking,'' in \emph{Proceedings of the European Conference on Computer
  Vision}, 2014, pp. 688--703.

\bibitem{hirzer2011person}
M.~Hirzer, C.~Beleznai, P.~M. Roth, and H.~Bischof, ``Person re-identification
  by descriptive and discriminative classification,'' in \emph{Scandinavian
  Conference on Image Analysis}.\hskip 1em plus 0.5em minus 0.4em\relax
  Springer, 2011, pp. 91--102.

\bibitem{zheng2016person}
L.~Zheng, Y.~Yang, and A.~G. Hauptmann, ``Person re-identification: Past,
  present and future,'' \emph{arXiv:1610.02984}, 2016.

\bibitem{li2018diversity}
S.~Li, S.~Bak, P.~Carr, and X.~Wang, ``Diversity regularized spatiotemporal
  attention for video-based person re-identification,'' in \emph{Proceedings of
  the IEEE Conference on Computer Vision and Pattern Recognition}, 2018, pp.
  369--378.

\bibitem{deng2009imagenet}
J.~Deng, W.~Dong, R.~Socher, L.-J. Li, K.~Li, and L.~Fei-Fei, ``Imagenet: A
  large-scale hierarchical image database,'' in \emph{Proceedings of the IEEE
  Conference on Computer Vision and Pattern Recognition}, 2009, pp. 248--255.

\bibitem{bottou2010large}
L.~Bottou, ``Large-scale machine learning with stochastic gradient descent,''
  in \emph{Proceedings of COMPSTAT'2010}.\hskip 1em plus 0.5em minus
  0.4em\relax Springer, 2010, pp. 177--186.

\bibitem{wu2018exploit}
Y.~Wu, Y.~Lin, X.~Dong, Y.~Yan, W.~Ouyang, and Y.~Yang, ``Exploit the unknown
  gradually: One-shot video-based person re-identification by stepwise
  learning,'' in \emph{Proceedings of the IEEE Conference on Computer Vision
  and Pattern Recognition}, 2018, pp. 5177--5186.

\bibitem{zhao2019attribute}
Y.~Zhao, X.~Shen, Z.~Jin, H.~Lu, and X.~Hua, ``Attribute-driven feature
  disentangling and temporal aggregation for video person re-identification,''
  in \emph{Proceedings of the IEEE Conference on Computer Vision and Pattern
  Recognition}, 2019, pp. 4913--4922.

\bibitem{hou2019vrstc}
R.~Hou, B.~Ma, H.~Chang, X.~Gu, S.~Shan, and X.~Chen, ``Vrstc: Occlusion-free
  video person re-identification,'' in \emph{Proceedings of the IEEE Conference
  on Computer Vision and Pattern Recognition}, 2019, pp. 7183--7192.

\bibitem{li2019global}
J.~Li, J.~Wang, Q.~Tian, W.~Gao, and S.~Zhang, ``Global-local temporal
  representations for video person re-identification,'' in \emph{Proceedings of
  the IEEE International Conference on Computer Vision}, 2019, pp. 3958--3967.

\bibitem{Liu2019SpatiallyAT}
C.-T. Liu, C.-W. Wu, Y.-C.~F. Wang, and S.-Y. Chien, ``Spatially and temporally
  efficient non-local attention network for video-based person
  re-identification,'' \emph{arXiv:1908.01683}, 2019.

\bibitem{yang2020spatial}
J.~Yang, W.-S. Zheng, Q.~Yang, Y.-C. Chen, and Q.~Tian, ``Spatial-temporal
  graph convolutional network for video-based person re-identification,'' in
  \emph{Proceedings of the IEEE Conference on Computer Vision and Pattern
  Recognition}, 2020, pp. 3289--3299.

\bibitem{yan2020learning}
Y.~Yan, J.~Qin, J.~Chen, L.~Liu, F.~Zhu, Y.~Tai, and L.~Shao, ``Learning
  multi-granular hypergraphs for video-based person re-identification,'' in
  \emph{Proceedings of the IEEE Conference on Computer Vision and Pattern
  Recognition}, 2020, pp. 2899--2908.

\bibitem{jiang2021ssn3d}
X.~Jiang, Y.~Qiao, J.~Yan, Q.~Li, W.~Zheng, and D.~Chen, ``Ssn3d:
  Self-separated network to align parts for 3d convolution in video person
  re-identification,'' in \emph{Proceedings of the AAAI Conference on
  Artificial Intelligence}, 2021, pp. 1691--1699.

\bibitem{liu2021spatial}
J.~Liu, Z.-J. Zha, W.~Wu, K.~Zheng, and Q.~Sun, ``Spatial-temporal correlation
  and topology learning for person re-identification in videos,'' in
  \emph{Proceedings of the IEEE Conference on Computer Vision and Pattern
  Recognition}, 2021, pp. 4370--4379.

\bibitem{hou2021bicnet}
R.~Hou, H.~Chang, B.~Ma, R.~Huang, and S.~Shan, ``Bicnet-tks: Learning
  efficient spatial-temporal representation for video person
  re-identification,'' in \emph{Proceedings of the IEEE Conference on Computer
  Vision and Pattern Recognition}, 2021, pp. 2014--2023.

\bibitem{aich2021spatio}
A.~Aich, M.~Zheng, S.~Karanam, T.~Chen, A.~K. Roy-Chowdhury, and Z.~Wu,
  ``Spatio-temporal representation factorization for video-based person
  re-identification,'' in \emph{Proceedings of the IEEE International
  Conference on Computer Vision}, 2021, pp. 152--162.

\bibitem{eom2021video}
C.~Eom, G.~Lee, J.~Lee, and B.~Ham, ``Video-based person re-identification with
  spatial and temporal memory networks,'' in \emph{Proceedings of the IEEE
  International Conference on Computer Vision}, 2021, pp. 12\,036--12\,045.

\bibitem{zhou2017see}
Z.~Zhou, Y.~Huang, W.~Wang, L.~Wang, and T.~Tan, ``See the forest for the
  trees: Joint spatial and temporal recurrent neural networks for video-based
  person re-identification,'' in \emph{Proceedings of the IEEE Conference on
  Computer Vision and Pattern Recognition}, 2017, pp. 4747--4756.

\bibitem{xu2017jointly}
S.~Xu, Y.~Cheng, K.~Gu, Y.~Yang, S.~Chang, and P.~Zhou, ``Jointly attentive
  spatial-temporal pooling networks for video-based person re-identification,''
  in \emph{Proceedings of the IEEE International Conference on Computer
  Vision}, 2017, pp. 4733--4742.

\bibitem{chen2018video}
D.~Chen, H.~Li, T.~Xiao, S.~Yi, and X.~Wang, ``Video person re-identification
  with competitive snippet-similarity aggregation and co-attentive snippet
  embedding,'' in \emph{Proceedings of the IEEE Conference on Computer Vision
  and Pattern Recognition}, 2018, pp. 1169--1178.

\bibitem{liu2019spatial}
Y.~Liu, Z.~Yuan, W.~Zhou, and H.~Li, ``Spatial and temporal mutual promotion
  for video-based person re-identification,'' in \emph{Proceedings of the AAAI
  Conference on Artificial Intelligence}, 2019, pp. 8786--8793.

\bibitem{chen2020frame}
Z.~Chen, Z.~Zhou, J.~Huang, P.~Zhang, and B.~Li, ``Frame-guided region-aligned
  representation for video person re-identification,'' in \emph{Proceedings of
  the AAAI Conference on Artificial Intelligence}, 2020, pp. 10\,591--10\,598.

\bibitem{li2020appearance}
S.~Li, H.~Yu, and H.~Hu, ``Appearance and motion enhancement for video-based
  person re-identification,'' in \emph{Proceedings of the AAAI Conference on
  Artificial Intelligence}, 2020, pp. 11\,394--11\,401.

\bibitem{sandler2018mobilenetv2}
M.~Sandler, A.~Howard, M.~Zhu, A.~Zhmoginov, and L.-C. Chen, ``Mobilenetv2:
  Inverted residuals and linear bottlenecks,'' in \emph{Proceedings of the IEEE
  Conference on Computer Vision and Pattern Recognition}, 2018, pp. 4510--4520.

\bibitem{huang2017densely}
G.~Huang, Z.~Liu, L.~Van Der~Maaten, and K.~Q. Weinberger, ``Densely connected
  convolutional networks,'' in \emph{Proceedings of the IEEE Conference on
  Computer Vision and Pattern Recognition}, 2017, pp. 4700--4708.

\bibitem{touvron2021training}
H.~Touvron, M.~Cord, M.~Douze, F.~Massa, A.~Sablayrolles, and H.~J{\'e}gou,
  ``Training data-efficient image transformers \& distillation through
  attention,'' in \emph{International Conference on Learning
  Representations}.\hskip 1em plus 0.5em minus 0.4em\relax PMLR, 2021, pp.
  10\,347--10\,357.

\bibitem{dai2019video}
J.~Dai, P.~Zhang, D.~Wang, H.~Lu, and H.~Wang, ``Video person re-identification
  by temporal residual learning,'' \emph{IEEE Transactions on Image
  Processing}, vol.~28, pp. 1366--1377, 2019.

\bibitem{selvaraju2017grad}
R.~R. Selvaraju, M.~Cogswell, A.~Das, R.~Vedantam, D.~Parikh, and D.~Batra,
  ``Grad-cam: Visual explanations from deep networks via gradient-based
  localization,'' in \emph{Proceedings of the IEEE International Conference on
  Computer Vision}, 2017, pp. 618--626.

\end{thebibliography}

\end{document}